\documentclass[10pt,twocolumn,letterpaper]{article}

\usepackage[pagenumbers]{cvpr} 
\usepackage[dvipsnames]{xcolor}

\usepackage[accsupp]{axessibility}
\usepackage{array}
\usepackage{tabularx}
\usepackage[toc]{appendix}
\definecolor{cvprblue}{rgb}{0.21,0.49,0.74}
\usepackage[pagebackref,breaklinks,colorlinks,citecolor=cvprblue]{hyperref}
\usepackage[font={small}]{caption}
\usepackage{multirow}
\def\paperID{3608} 
\def\confName{CVPR}
\def\confYear{2024}

\usepackage{pifont} 
\newcommand{\cmark}{\ding{51}}
\newcommand{\xmark}{\ding{55}}

\definecolor{highlightblue}{HTML}{D9E2ED} 
\definecolor{highlightred}{HTML}{FFB2B2} 
\definecolor{highlightyellow}{HTML}{F8EBBE} 

\title{CosmicMan: A Text-to-Image Foundation Model for Humans}

\author{
Shikai Li{$^*$}, \quad Jianglin Fu{$^*$},  \quad Kaiyuan Liu{$^*$}, \quad Wentao Wang{$^*$}, \quad Kwan-Yee Lin\textsuperscript{$\dagger$}, \quad Wayne Wu\textsuperscript{$\dagger$}  \\
Shanghai AI Laboratory \\
{\tt\small \{lishikai, fujianglin, wangwentao\}@pjlab.org.cn} \\ {\tt\small 1154864382@mail.dlut.edu.cn} \quad {\tt\small linjunyi9335@gmail.com} \quad {\tt\small wuwenyan0503@gmail.com}
}

\begin{document}

\twocolumn[{
            \renewcommand\twocolumn[1][]{#1}
            \maketitle
            \begin{center}
                \centering
                \includegraphics[width=0.99\textwidth]{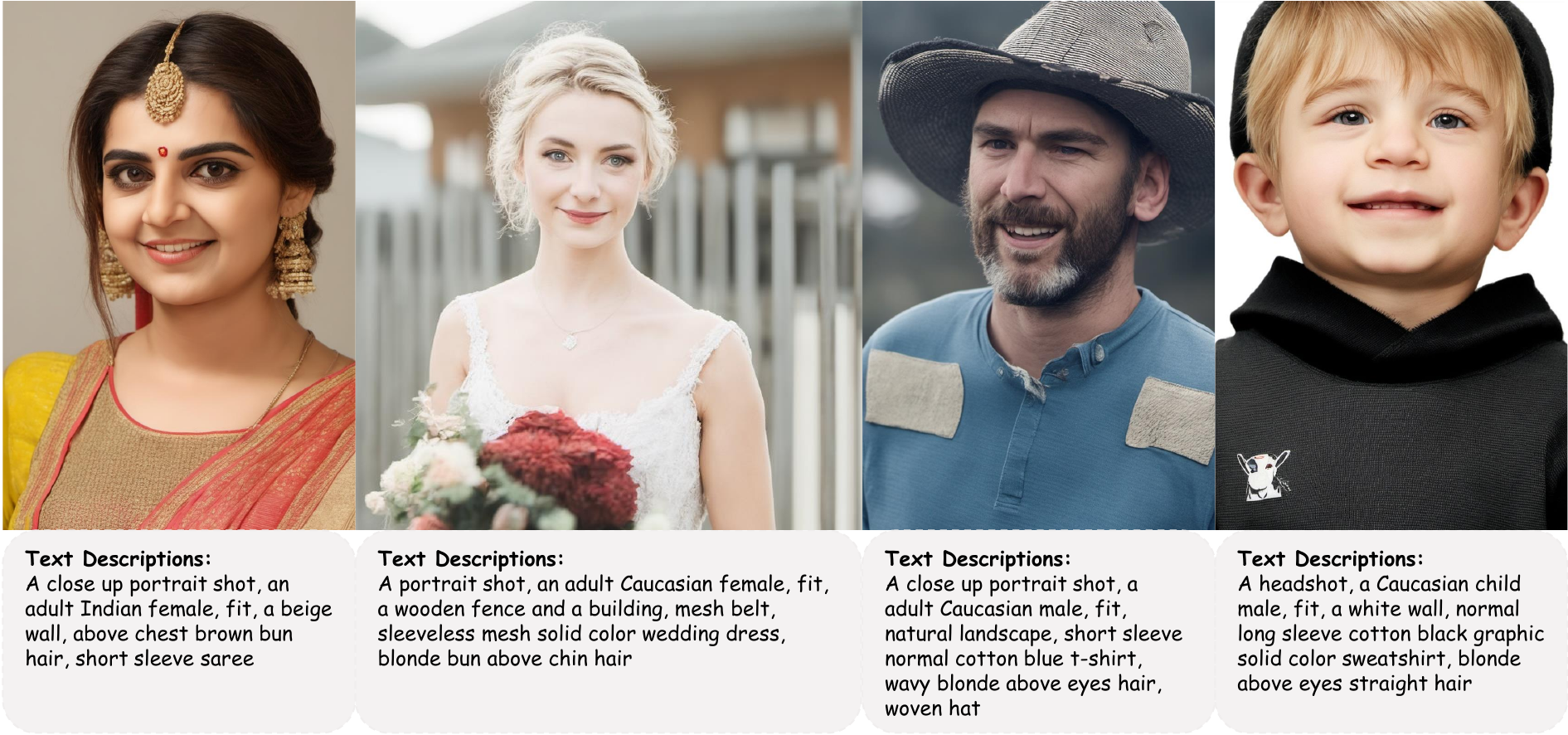}
                \captionof{figure}{\textbf{CosmicMan.} High-fidelity images generated by our proposed human-specialized text-to-image foundation model CosmicMan. The results are with meticulous appearance, reasonable structure, and precise text-image alignment with detailed dense descriptions. 
                More results are presented in the project page: \url{https://cosmicman-cvpr2024.github.io/}.
                }
                \label{fig:teaser}
            \end{center}
        }
        ]

\def\thefootnote{*}\footnotetext{Joint first authors.}
\def\thefootnote{\arabic{footnote}} 
\def\thefootnote{$\dagger$}\footnotetext{Equal advising.}
\def\thefootnote{\arabic{footnote}}

\begin{abstract}
We present \textbf{CosmicMan}, a text-to-image foundation model specialized for generating high-fidelity human images. Unlike current general-purpose foundation models that are stuck in the dilemma of inferior quality and text-image misalignment for humans, CosmicMan enables generating photo-realistic human images with meticulous appearance, reasonable structure, and precise text-image alignment with detailed dense descriptions.

At the heart of CosmicMan's success are the new reflections and perspectives on data and models:
$(1)$ We found that data quality and a scalable data production flow are essential for the final results from trained models. Hence, 
we propose a new data production paradigm, \textbf{Annotate Anyone}, which serves as a perpetual data flywheel to produce high-quality data with accurate yet cost-effective annotations over time. Based on this, we constructed a large-scale dataset, \textbf{CosmicMan-HQ 1.0}, with $6$ Million high-quality real-world human images in a mean resolution of $1488\times 1255$, and attached with precise text annotations deriving from $115$ Million attributes in diverse granularities.
$(2)$ We argue that a text-to-image foundation model specialized for humans must be pragmatic -- easy to integrate into down-streaming tasks while effective in producing high-quality human images. Hence, 
we propose to model the relationship between dense text descriptions and image pixels in a decomposed manner, and present {\textbf{D}}ecomposed-{\textbf{A}}ttention-{\textbf{R}}efocus{\textbf{ing}} (\textbf{Daring}) training framework.
It 
seamlessly decomposes the cross-attention features in existing text-to-image diffusion model, and enforces attention refocusing without adding extra modules. Through Daring, we show that explicitly discretizing continuous text space into several basic groups that align with human body structure is the key to tackling the misalignment problem in a breeze.

\end{abstract}    
\section{Introduction}
\label{sec:intro}

Text-to-image foundation models, \eg, Stable Diffusion (SD)~\cite{sd}, Imagen~\cite{imagen}, and DALLE~\cite{dalle}, have made groundbreaking contributions in the realm of Computer Vision and Graphics. These models, fueled by expansive image-text datasets~\cite{laion5b, coyo} and advanced generative algorithms~\cite{vqvae, taming, ddpm}, possess the capability to create images of remarkable quality and details. These models, underpinned by robust prior knowledge, have significantly enhanced a wide array of downstream tasks. Notable examples include DreamBooth~\cite{dreambooth} and ControlNet~\cite{controlnet} in 2D image generation, alongside DreamFusion~\cite{dreamfusion} and Zero-1-to-3~\cite{zero123} in 3D object creation. Despite these advances, a critical gap remains within the sphere of \textit{human-centric} content generation -- the absence of a specialized text-to-image foundation model that serves as a cornerstone for tasks on human subjects.

In prior research, tasks related to human-centric content generation, like 2D human generation/editing~\cite{sghuman, jiang2022text2human, fashiontex} and 3D human generation/reconstruction~\cite{4k4d, pifu, 3dhumangan, ortho}, have typically progressed in isolation, each relying on its in-domain data.
These methods, however, faced a key limitation: the datasets used were often narrow in diversity~\cite{dnarendering}, 
exhibited biased distributions~\cite{sghuman}, 
or lacked quality~\cite{humansd}. 
Achieving \textit{generalization} across a broad range of identities, appearances, and geometries in real-world applications has been challenging within this framework. Nevertheless, the emergence of text-to-image foundation models, which have excelled in the general-purpose arena, offers a promising new direction to revolutionize human-centric content generation with enhanced generalization capabilities. 

The pivotal question then arises: \textit{How to obtain a text-to-image foundation model for humans?}
By analyzing the essential demands of a human-specialized foundation model, we identify three critical elements necessary for such a model: 
1) \textbf{High-Quality Data.} To train a foundation model that will be used in downstream tasks to generate high-quality content, the raw data quality is critical. The raw data quality encompasses not only the volume but also the image quality and diversity, as well as the precision, granularity, and comprehensiveness of annotations. While large-scale datasets featuring text-image pairs (\textit{e.g.,} LAION-5B~\cite{laion5b}, and COYO-700M~\cite{coyo}) have advanced general-purpose foundation models, they often deviate from accurately representing real-world human distributions, suffering from jagged image quality and a mass of annotation noise.
2) \textbf{Scalable Data Production.} A foundation model that generalizes effectively must evolve in sync with the growth of real-world data. Given the vast amount of training data that is requisite and the rapid pace at which it expands in human contexts, developing a scalable data production process is imperative -- being updated over time and cost-effective. Traditional methods often incur high costs due to manual annotation~\cite{mscoco, textcaps, nocaps, flickr30k}, or suffer from accuracy issues when using automated labeling~\cite{cc}. Moreover, the reliance on static datasets limits their ability to adjust according to dynamic real-world data distributions.
3) \textbf{Pragmatic Model.} A foundation model designed for humans should be straightforward to integrate into downstream tasks, requiring minimal customization of its architecture. In addition, given the complexity of human anatomy, the ability of a model to generate high-quality outputs is also essential -- the outputs should guarantee realistic structures and precise text-image alignment, especially capturing detailed dense concepts attached to humans. Existing models, whether closed-source
like MidJourney~\cite{midjourney}, DALLE~\cite{dalle}, or those struggling with high-fidelity human generation such as
SD~\cite{sd} and SDXL~\cite{sdxl}, highlight the need for a versatile, high-quality model tailored for human-centric applications.

We present \textbf{CosmicMan} -- a holistic solution of text-to-image foundation model for humans.
We first introduce a new data production paradigm \textit{Annotate Anyone} by human-AI cooperation, which can produce flowing, high-quality yet cost-effective data continuously. Annotate Anyone consists of two main stages: \textit{Flowing Data Sourcing} to get a flowing data pool with irrigated high-quality human images from academic datasets and Internet, 
and \textit{Human-in-the-loop Data Annotation} to iteratively refine the labeling quality of the data in the pool at a fairly low cost. Then, a data flywheel is constructed to produce vast amounts of data in a dynamic, up-to-date, and economical manner, which is well-adaptive in the age of large-scale foundation models.

By running Annotate Anyone, we constructed a large-scale, high-quality dataset, \textit{CosmicMan-HQ 1.0}, which currently included $6$ million human images with a mean resolution of $1488\times1255$. It includes rich annotations with high precision -- $115$ million attributes, texts, bounding boxes, keypoints, human parsings, and rich meta information. 
Empowered by Annotate Anyone, CosmicMan-HQ continues to grow rapidly. Future versions of CosmicMan-HQ will support the perpetual update of foundation models with growing real-world data, facilitating long-term research in human-centric generation. 

Finally, based on CosmicMan-HQ, we provide a human-specialized foundation model to support the human-centric content generation tasks. To ensure the easy of use of the proposed model, we construct our model by tailoring SD with minimum modification. Concretely, we introduce {\textit{D}}ecomposed-{\textit{A}}ttention-{\textit{R}}efocus{\textit{ing}} (\textit{Daring}), a training framework that is rooted in SD without adding extra modules. In virtue of the nature of the proposed dataset, the key insight of the framework is explicitly discretizing dense descriptions into a fixed number of groups related to human body structure. Based on this, we could decompose the cross-attention feature maps according to the groups and enforce the network learning attention refocusing at the group level. This target could be achieved by adding a new loss supervised on cross-attention maps, 
called \textit{HOLA} (short for {\textit{H}uman Body and {\textit{O}utfit} Guided {\textit{L}oss} for {\textit{A}lignment}).

In experiments, we demonstrate superior image quality and text-image alignment by comparing our models to state-of-the-art foundation models. Then, we conduct extensive ablation studies to show the effectiveness of our designs in data production and model training. Finally, we show the pragmaticality and potential of our human-specialized foundation model with applications in 2D and 3D  generation.

\section{Related Work}
\label{sec:relatedwork}

\subsection{Text-to-Image Foundation Models} 
Advancements in data volume and model design have led text-to-image foundation models to produce high-fidelity images that follow the text instructions.
DALLE~\cite{dalle}, which pioneered zero-shot text-to-image generation, autoregressively modeling text and image tokens in a unified data stream. 
Its successors~\cite{dalle2, dalle3} enhance performance through model design and improved captions. 
Imagen~\cite{imagen} utilizes a larger text encoder with better photo realism. 
Open-source models like DeepFloyd-IF~\cite{deepfloyd}, PixelArt-$\alpha$~\cite{pixelalpha}, and particularly SD~\cite{sd}, along with SDXL~\cite{sdxl}, have energized the community, accelerating various applications in downstream tasks. 
In 2D content generation, innovations like ControlNet~\cite{controlnet} and T2I-Adapter~\cite{t2i-adapter} have emerged, and in 3D, models like Zero-1-to-3~\cite{zero123} and DreamFusion~\cite{dreamfusion} leverage SD to create high-quality 3D objects.
However, these foundation models are geared towards the general-purpose domain, which falls short in generating humans due to their tendency to overlook the nuances and complexities of human anatomy and attire. 
There is still a gap for a human-specialized text-to-image foundation model to boost downstream human content generation.

\subsection{Text-Driven Human Image Generation}
Previous research~\cite{ladi, jiang2022text2human, fashiontex, fice} primarily focused on fashion-domain data and has achieved high-fidelity human generation and editing with control over text. For instance, Text2Human~\cite{jiang2022text2human} employs a two-stage framework using VQ-VAE~\cite{vqvae} to transfer human pose into human parsing with cloth shape and generates human images with texture description. 
FashionTex~\cite{fashiontex} offers text and texture-based control for virtual try-on by leveraging the pretrained generative models~\cite{sghuman}. 
However, the extremely biased and insufficient training data, hinder the diversity of generated images from these methods. Meanwhile, some approaches~\cite{humansd, controlnet} leverage text-to-image foundation models~\cite{sd} to create more diverse human images with additional conditions (\textit{e.g.,} skeletons and normal maps). HumanSD~\cite{humansd} introduces a skeleton-guided diffusion model trained on a diverse dataset that enhances the accuracy of pose control. However, these methods are used for controllable human image generation.
In contrast, CosmicMan stands out as a foundation model by producing high-quality and diverse human images without relying on spatial conditions in the inference phase. 

\subsection{Text-Image Alignment for Dense Concepts}
Early text-to-image models, often trained on short text captions, struggle to encapsulate dense concepts present in longer descriptions. The dense concepts, as discussed in many text-to-image benchmarks~\cite{compbench,geneval}, include multiple objects, attributes, and spatial relationships that describe the image from different granularities and perspectives. 
The challenge lies in generating each element and accurately depicting their interrelations within long descriptions.
Recently, the training-free methods~\cite{p2p, attend_and_excite, structurediffusion, decompose_and_realign, diffcloth, li2023divide, linguistic} found that the cross-attention mechanism plays a pivotal role in text-to-image alignment. Prompt-to-Prompt~\cite{p2p} first reveals that the cross-attention map governs the semantics of the output. Subsequent methods~\cite{attend_and_excite, diffcloth, decompose_and_realign, li2023divide} are proposed to employ the gradients of the well-designed loss to update the latent feature along the diffusion process. Additionally, FastComposer~\cite{xiao2023fastcomposer} applies the supervision of cross-attention maps during training. 
However, intuitively applying this to human images becomes more complex, as dense captions for humans often cluster in a small image region, as shown in Fig.~\ref{fig:data_samplings_parsings}. 
Thus, we propose a training framework that utilizes human-specific prior on arrangement relationships to supervise the cross-attention maps with decomposed text-human image data, which further improves the text-image alignment on dense concepts.

\section{Annotate Anyone -- A Data Flywheel}
To enable the learning of the human-specialized foundation model, we propose a human-AI cooperation paradigm for data production named \textbf{Annotate Anyone}. It combines the strengths of AI and human expertise to build a continuously expandable dataset CosmicMan-HQ with rich annotations. 
In this section, we will first introduce Annotate Anyone by comparing to previous data production paradigms (Sec.~\ref{sec:data_paradigm}). Then, we will elaborate on the procedures of Annotate Anyone in data sourcing and annotation (Sec.~\ref{sec:annotateanyone}). Finally, we will analyze the statistics of CosmicMan-HQ 1.0 dataset produced by Annotate Anyone, and show its superiority to existing datasets (Sec.~\ref{sec:datastats}).


\subsection{Data Production by Human-AI Cooperation}
\label{sec:data_paradigm}
\begin{figure}
    \centering
    \includegraphics[width=1\linewidth]{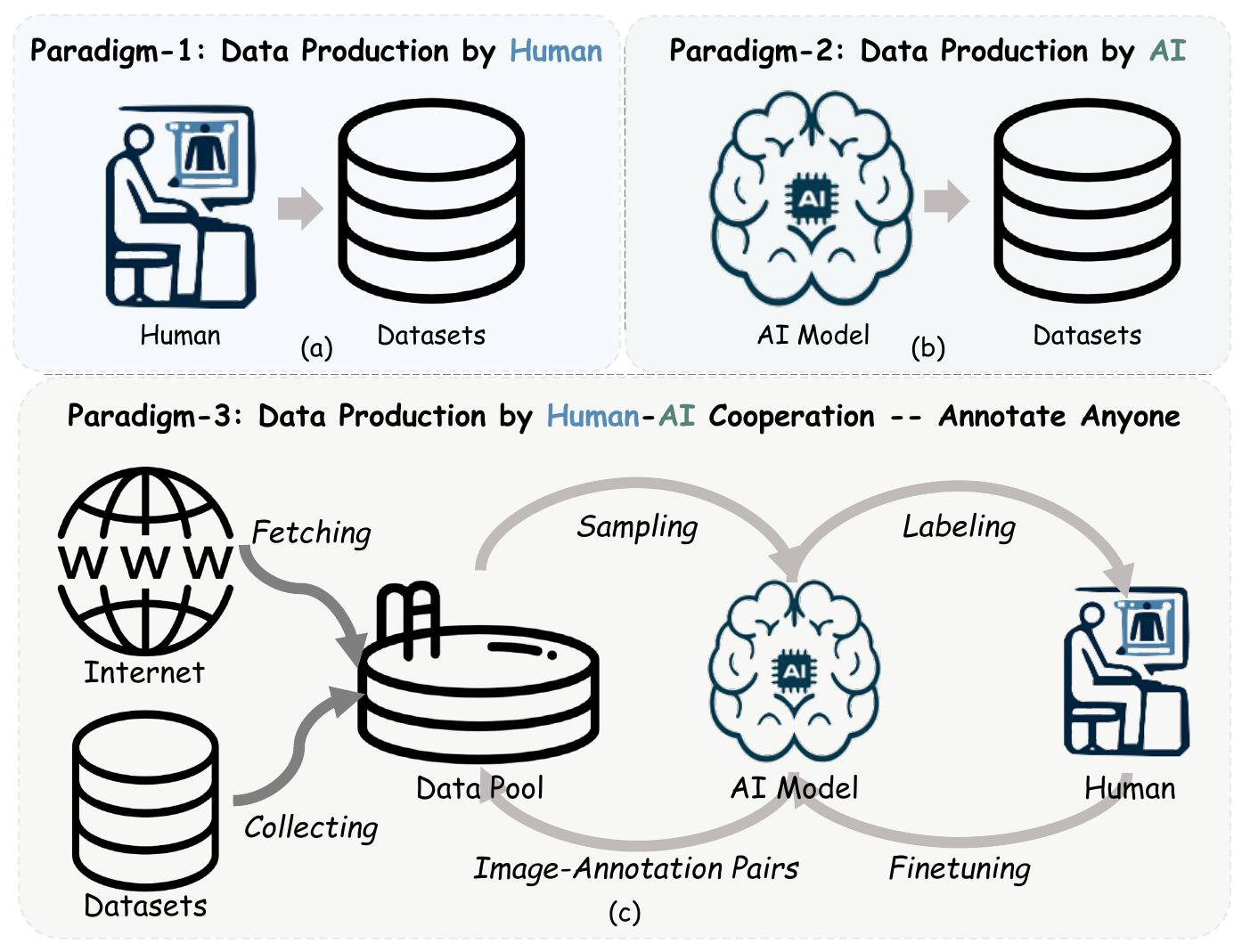}
    \caption{\textbf{Data Production Paradigm.} (a) Data production by humans and (b) data production by AI. (c) Our proposed new data production paradigm by Human-AI cooperation, named Annotate Anyone. It serves as a \textit{data flywheel} to produce dynamic up-to-date high-quality data at a low cost.}
    \label{fig:data_production_paradigm}
\end{figure}

To construct large-scale image datasets with labels, there are mainly two paradigms -- by humans or by AI. Data production by humans (as depicted in Fig.~\ref{fig:data_production_paradigm} (a)) needs human annotators to manually label images one by one~\cite{imagenet,mscoco,ffhqtext}, 
which suffers from its high cost and thus is hard to scale up to support the recent development of large foundation models.
On the other hand, data production by AI (as depicted in Fig.~\ref{fig:data_production_paradigm} (b)) uses off-the-shelf models to get labels for free~\cite{dalle3,pixelalpha}. 
Although this paradigm dramatically reduces costs and is easy to scale up, it is notorious for its noisy, jagged, and coarse labeling results. Moreover, both of these paradigms rely on fixed datasets for labeling, which results in limited diversity and severe bias versus real-world data.
To train a large foundation model, a huge quantity of data, high-precise and fine-grained labeling, and real-world distribution are all indispensable. Thus, these paradigms are especially knotty to adapt to the human domain.

To this end, we propose a new data production paradigm by \textit{human-AI cooperation} named Annotate Anyone (as shown in Fig.~\ref{fig:data_production_paradigm} (c)). Compared to data production by humans and AI, Annotate Anyone pivots on two characteristics: 1) flowing data, and 2) human-in-the-loop annotation.
Flowing data is sourced from two origins: existing published datasets and the Internet. By collecting data from existing published datasets, such as SHHQ~\cite{sghuman} and LAION-5B~\cite{laion5b}, 
we can upcycle them to match the qualifications of a high-quality human dataset. By fetching data from the Internet, we can obtain the massive data produced by human beings every second. Our data sourcing system is always on call to run when the data quantity triggers the lower-bound threshold. Thus, our data pool is continuously \textit{flowing and refreshed}, which distinguishes it from previous paradigms.
Human-in-the-loop annotation coordinates three entities: data pool, AI, and human annotators, to work in a circle. By labeling a small quantity of data with the greatest necessity, human annotators and AI models cooperate to iteratively improve the quality of data annotation. Consequently, data in the pool will have progressively better annotation quality at a minimal cost.

With Annotate Anyone, we construct a \textbf{data flywheel} to enable a dynamic up-to-date production of high-quality data. Consequently, just by running the Annotate Anyone workflow, we can construct a large-scale human dataset with continuously accumulated data with great ease.

\subsection{Procedure of Annotate Anyone}
\label{sec:annotateanyone}

The procedure of Annotate Anyone consists of two main parts: 1) flowing data sourcing to obtain and filter high-quality data from both academic datasets and the Internet continually, and 2) human-in-the-loop data annotation to get precise yet cost-effective labels covering detailed dense concepts and ensuring high-quality annotations.

\subsubsection{Flowing Data Sourcing}

We first source images from various origins to ensure massive quantity and catch the real-world distribution. Then, we design a data filter to eliminate unbefitting images for human content generation tasks.

\noindent
\textbf{Data Origins.}
\label{sec:datacollection}
We start with three {\textit{academic}} datasets to recycle existing data resources: LAION-5B~\cite{laion5b}, SHHQ~\cite{sghuman}, and DeepFashion~\cite{deepfashion}. LAION-5B is a renowned collection of massive images shared online, while the other two datasets are smaller in scale and diversity but meticulously curated to ensure high quality.
Then, we initiate $128$ parallel processes in $32$ CPU servers, monitoring a wide spectrum of APIs on the Internet, including Flickr~\cite{FlickrAPI}, Unsplash~\cite{UnsplashAPI}, Pixabay~\cite{PixabayAPI}, \etc. These APIs give access to a vast collection of growing and diverse images, rendering a real-world distribution.

\noindent \textbf{Data Filtering.}
\label{sec:datafiltering}
The current data pool exhibits a broad distribution, but high-resolution human images are not the primary constituent. We use a set of data filtering strategies to distill a high-quality human-centric subset, including fake-people detection, image quality assessment, and so on. 
To remove fake-people images (\eg, cartoon characters, mannequin models, and generated images), we fine-tuned Eva-CLIP~\cite{evaclip} with Human-Art~\cite{humanart} and sets of fake images. The fine-tuned model with $91\%$ accuracy is used to detect images containing fake people. Next, image quality assessment metrics (LIQE~\cite{liqe}, IFQA~\cite{ifqa}, and HPSv2~\cite{hpsv2}) are applied, estimating the image quality on both face and global levels, to streamline the data pool further. 
Further, we utilize YOLOv7~\cite{yolov7} to filter out images without humans or those containing more than one individual. Images where the largest detected face is smaller than $224\times224$ or where the image is smaller than $640\times1280$ are removed as well. See more details in the supplementary material.

\subsubsection{Human-in-the-loop Data Annotation}
\label{sec:human-in-the-loop}

Having a data pool with diverse and high-quality human images, the next step is to possess precise, fine-grained yet cost-effective annotations for the images. We propose a human-in-the-loop data annotation workflow to iteratively refine the labeling quality of the data in the pool. 
Below, we first introduce the annotation iteration and then discuss the label protocol we use to describe humans in a detailed way.

\begin{figure}[t!]
    \centering
    \includegraphics[width=1\linewidth]{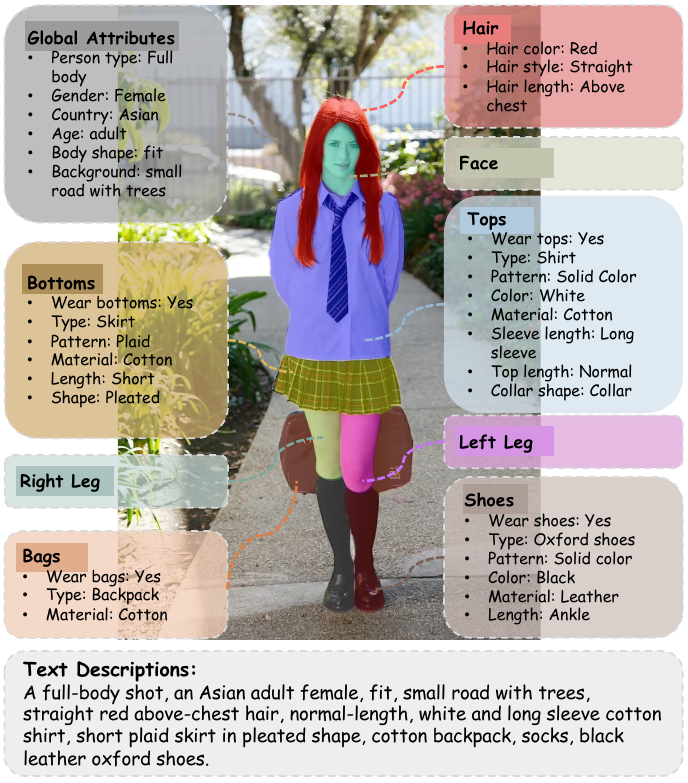}
    \caption{\textbf{Parsing Examples from CosmicMan-HQ.} The parsing results of sampled image in our dataset, along with detailed labels for each part. Text descriptions are obtained from labels.
    } 
    \label{fig:data_samplings_parsings}
\end{figure}

\begin{table*}[t!]
\caption{\textbf{Dataset Comparison.} The statistical comparison between publicly available human-related datasets and CosmicMan-HQ 1.0. ``Common Scale'' refers to the dataset that includes images captured at common scales, such as full-body shots, portrait photos, and half-body shots. ``HP'' and ``Aes'' refer to Human Parsing maps and Aesthetic scores. }
\centering
\label{tab:datasetcomp}
\resizebox{1\linewidth}{!}{
\begin{tabular}{cccccccccccccc}  
    \toprule
    & \multicolumn{3}{c|}{Data Quantity} & \multicolumn{2}{c|}{Imaging Quality} & \multicolumn{7}{c|}{Annotation} & \multirow{3}{*}{Domain} 
    \\ \cline{2-13} 
    
    & \begin{tabular}[c]{@{}c@{}}Total \\ Image \#\end{tabular} 
    & \multicolumn{1}{c}{\begin{tabular}[c]{@{}c@{}}Mean \\ Resolution\end{tabular}} 
    & \multicolumn{1}{c|}{\begin{tabular}[c]{@{}c@{}}Common \\ Scale\end{tabular}} 
    & Global$\uparrow$ & \multicolumn{1}{c|}{Face$\uparrow$} & Cat \# & Attr \# & Text & Bbox & Kpts & HP & \multicolumn{1}{c|}{Aes} & \\ 
    \midrule  
    Human-Art~\cite{humanart} & $50K$  & $1115\times1287$ & \textcolor[HTML]{A5B592}{\cmark} &  3.42 & 2.87 & - & - & \textcolor[HTML]{A5B592}{\cmark} & \textcolor[HTML]{A5B592}{\cmark} & \textcolor[HTML]{A5B592}{\cmark}  &   \textcolor[HTML]{D092A7}{\xmark} & \textcolor[HTML]{D092A7}{\xmark}& Real world \& AI \\ 
    DF-MM~\cite{jiang2022text2human} & $44K$ & $750\times1101$ & \textcolor[HTML]{D092A7}{\xmark} & 4.64  & 3.38 & 18 & $587K$ & \textcolor[HTML]{A5B592}{\cmark} & \textcolor[HTML]{A5B592}{\cmark} &  \textcolor[HTML]{D092A7}{\xmark} & \textcolor[HTML]{A5B592}{\cmark} &\textcolor[HTML]{D092A7}{\xmark} & Fashion\\ 
    LAION-Human~\cite{humansd} & $1M$ & $688\times650$ & \textcolor[HTML]{A5B592}{\cmark} & 4.20  & 2.66 & - & - & \textcolor[HTML]{A5B592}{\cmark} & \textcolor[HTML]{D092A7}{\xmark} & \textcolor[HTML]{D092A7}{\xmark} &\textcolor[HTML]{D092A7}{\xmark} & \textcolor[HTML]{A5B592}{\cmark} & Real world\\ 
    SHHQ 1.0~\cite{sghuman} & $40K$ & $1024\times512$ & \textcolor[HTML]{D092A7}{\xmark} & 4.23 & 2.13 & - & - & \textcolor[HTML]{D092A7}{\xmark} & \textcolor[HTML]{D092A7}{\xmark} & \textcolor[HTML]{A5B592}{\cmark} & \textcolor[HTML]{A5B592}{\cmark} & \textcolor[HTML]{D092A7}{\xmark} & Fashion \\ 
    \midrule
    CosmicMan-HQ 1.0 & $6M$ & $1488\times1255$ & \textcolor[HTML]{A5B592}{\cmark} &4.37 & 3.37 & 70 & $115M$ & \textcolor[HTML]{A5B592}{\cmark} & \textcolor[HTML]{A5B592}{\cmark} & \textcolor[HTML]{A5B592}{\cmark} & \textcolor[HTML]{A5B592}{\cmark} & \textcolor[HTML]{A5B592}{\cmark} & Real world \\ 
    \bottomrule 
\end{tabular}
}
\end{table*}

\noindent
\textbf{Annotation Iteration.} As shown in Fig.~\ref{fig:data_production_paradigm} (c), the iterations start from sampling an image set $I_i$ from the data pool and end up with putting all image-annotation pairs $(I,A)$ back to the data pool. We set an evaluation set $I_e$ with ground truth. In each iteration, $I_e$ is used to determine the categories that need to be labeled by human annotators, and $I_i$ will be partially labeled with the selected categories. Then, $I_i$ is used to finetune the AI model. The finetuned AI model is evaluated on $I_e$ to determine whether to continue or stop the iterations. Finally, a well-finetuned AI model is used to get image-annotation pair $(I,A)$ for all data in the pool. Specifically, inspired by methods ~\cite{dalle3,pixelalpha} 
that use the Vision-Language Model (VLM) to perform image captioning tasks, we leverage a pretrained InstructBLIP~\cite{instructblip} as our AI model in the iteration. Please refer to the supplementary for the pseudo-code of the annotation iteration.

The pivotal mechanism to implement the high-precise yet low-cost annotation is the trigger of human annotation. During the initial iteration, the annotation team labels all categories based on $70$ questions.  
We observed that the accuracy of the predicted labels follows a real-world distribution, exhibiting a long-tail distribution. For the head categories, such as age and gender, the pretrained AI model already proficients in the prediction. 
Thus, in subsequent iterations, human annotators focus on tail categories, and categories with an accuracy above $85\%$ will no longer be manually labeled. 
Our iterative process significantly improved the VLM model's overall accuracy by at least $30\%$ compared to the pretrained model. 
Moreover, the progressive reduction of labeled annotations during the iterations resulted in only $1\%$ compared to full manual labeling. 
Please refer to the supplementary materials for experimental details.

\noindent
\textbf{Label Protocol.} 
\label{sec:label_protocol}
To systematically describe a human image comprehensively, we design a label protocol that leverages the human parsing model SCHP~\cite{schp} to break down an image into $18$ fine parts, including background, face, upper clothing, and \etc. Each part has an average of 3 to 8 associated questions, resulting in a total of 70 questions that correspond to 70 categories. For example, ``top-sleeve length'' is one of the categories. The answers under each category form the detailed attributes in our dataset. For instance, the attributes associated with ``top - sleeve length'' include ``long sleeve, mid-sleeve, sleeveless'', and so on. As shown in Fig.~\ref{fig:data_samplings_parsings}, for every single clothing item, once we confirm the existence of a particular garment, we proceed to further inquire about deeper detailed questions, such as color, pattern, and material. Questions related to global attributes are also asked. By employing this label protocol, we obtain a rich set of attributes regarding the human image with each label corresponding to a specific spatial position. Please refer to supplementary materials to see our final labeling results with the comparison to others.

\subsection{CosmicMan-HQ 1.0 Dataset}
\label{sec:datastats}
By running Annotate Anyone, so far, the first version of the produced dataset CosmicMan-HQ 1.0 consists of $6M$ high-resolution, single-person images, along with corresponding rich annotations. Here we compare our dataset with representative human-centric datasets in terms of data quantity, imaging quality, and annotations.

As depicted in Tab.~\ref{tab:datasetcomp}, our dataset is the largest crafted human-centric dataset, six times larger than LAION-Human~\cite{humansd}. The mean resolution is $1488\times1255$, surpassing previous human-only datasets like DF-MM~\cite{jiang2022text2human} and SHHQ~\cite{sghuman} by a large margin. Our dataset possesses a diverse collection of human images, including full-body shots, headshots, half-body shots, and so on. In terms of image quality at both overall and face level, our dataset ranks second only to the fashion-focused DF-MM dataset, which predominantly contains professional studio images but with less diversity and a data amount. As for annotation, only DF-MM and ours provide manually labeled categories, but the former dataset is much smaller in data volume and category numbers. 
CosmicMan-HQ 1.0 provides $70$ categories and around $115M$ detailed attributes as detailed in the Label Protocol part of Sec.~\ref{sec:label_protocol}.

Highlighting our dataset's uniqueness, CosmicMan-HQ 1.0 distinguishes itself by providing an unparalleled wealth of diverse annotations, including $115M$ attributes, texts, bounding boxes, keypoints, human parsings, and rich meta information (web alternative texts, aesthetic scores, watermark scores, face/global quality scores, and camera EXIF parameters).

\section{Daring - The Training Framework}

We propose {\textbf{Daring}} ({\textbf{D}}ecomposed-{\textbf{A}}ttention-{\textbf{R}efocus{\textbf{ing}}), a training framework rooted in original Stable Diffusion (SD) with minimal modification. The framework is illustrated in Fig.~\ref{fig:method}. It enjoys three properties at the same time  --  friendly to computational costs, compatible with downstream tasks supported by SD, and robust in producing high-quality human images that align well with dense concepts. These come from two parts' design -- data discretion for decomposing text-human image data (Sec.~\ref{hp}), and a new loss aiming to improve the alignment with respect to the scale of the human body and outfits (Sec.~\ref{hola}). }

\subsection{Preliminaries}\label{pre}
We employ SD as the backbone model for its efficiency and widespread application in various downstream tasks. SD incorporates a variational autoencoder $\mathcal{E}$ to encode images $\mathbf{x}$ as latent variables $z$ in a compact latent space, and applies diffusion schema in the latent space, thereby facilitating the diffusion process and reducing the computational cost. The denoising network is optimized by minimizing the $L_2$ error between predicted noise $\epsilon_{\theta}$ and ground-truth noise $\epsilon\sim\mathcal{N}(\mathbf{0}, \mathbf{I})$:
\begin{equation}
\mathcal{L}_{{noise}}=\mathbb{E}_{z\sim\mathcal{E}(x),c,\epsilon,t}\left[||\epsilon-\epsilon_{\theta}(z_{t},t,c)||_{2}^{2}\right],
\end{equation}
where $z_t$ is the latent at time-step $t$, and $c$ is the condition information that can be instantiated by text input. 

The cross-attention layers are the hinge for textual information to play a role in influencing the updating of intermediate features. Specifically, a text prompt $\mathcal{P}$ is first transformed into a text embedding $c$ via a CLIP text encoder. The latent $z_t$ and text embedding are projected to form a query~$Q$ and keys~$K$. The cross-attention maps are computed to flatten textual information into spatial features:
\begin{equation}\label{attn}
M = \text{Softmax}(\frac{QK^{T}}{\sqrt{d}})
\end{equation}
where $d$ is the dimension of $Q$ and $K$ embeddings. The design works well when the text descriptions are short sparse captions. However, it can not handle text information with dense concepts, due to the lack of effective guidance to learn distinctive and precisely located features. 

\begin{figure}
    \centering
    \includegraphics[width=1\linewidth]{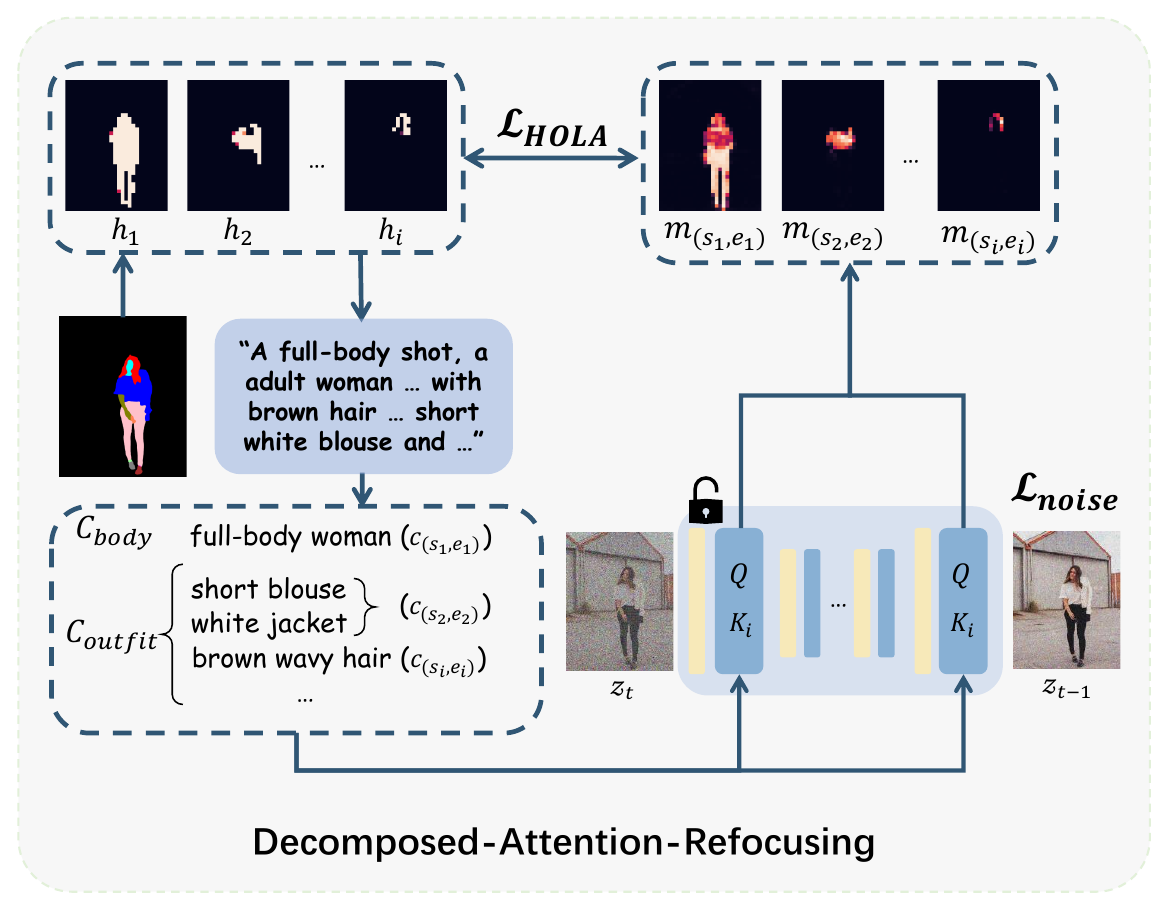}
    \caption{\textbf{Daring Training Framework.} It includes two parts: (1) data discretion for decomposing text-human data into fixed groups that obey human structure; (2) a new loss -- HOLA, to enforce the cross-attention features actively response in proper spatial region with respect to the scale of body structure and outfit arrangement.}
    \label{fig:method}
\end{figure}

\subsection{Data Discretization for Humans}\label{hp}
We argue that there is no necessity to optimize the latent code guided by cross-attention maps during inference or harm the original architecture design of SD with sophisticated modules, {\textit{as long as keys $K$ are decomposable and finite at the first place}}. This is doable and simple to achieve. Because for the human generation scenario, textual descriptions about a person always revolve around body structure and attachments. As humans are structured in nature, textual descriptions could be explicitly classified into fixed groups that correspond to body regions, no matter how many concepts are described.    

Thus, rather than directly utilizing nature language description, we propose a discretized textual prototype as illustrated in Fig.~\ref{fig:method} for the network to enable precise communication between tokens. The prototype defines the convention of classifying and arranging the concepts of text captions to a finite set, where all captions can be represented as $C = 	\left\{C_{body}, C_{outfit} \right\}$. The subset $C_{body}$ is for overall appearance, and the subset $C_{outfit}$ is for fine-grained attributes of outfits.

Concretely, as shown in Fig.~\ref{fig:method}, given a human data sample $\mathbf{x}$ in CosmicMan-HQ, we first reorganize human parsing maps into the semantic map sets $H=\{h_{i} \}_{i=1}^N$, where $N$ is the number of semantic masks and $h_{1}$ is the aggregation of all human parsing maps to differentiate human foreground with background. These masks are categorized into two levels -- $h_1$ lies in \textit{human-body-level} and the others belong to~\textit{outfit-level}. Then, we split the text captions with respect to $H$.  Specifically, $C_{body}= c_{(s_{1}, e_{1})}$ and $C_{outfit}= \{c_{(s_{2}, e_{2})},\ldots, c_{(s_{N}, e_{N})},c_{other}\}$.  $c_{(s_{n}, e_{n})}$ denotes the $n^{th}$ sub-caption group related to the semantic map $h_n$, and $s_{n}, e_{n}$ are the start and end indices of the concepts in the caption respectively. We gather the caption phrases without corresponding semantic masks as $c_{other}$, such as the caption for background. Note that, as our dataset naturally constructs annotation labels in a hierarchical manner, we can easily associate textual concepts with the semantic maps. For example, given a semantic map $h_2$ that represents the top clothing mask, we retrieve all the labels related to the top clothing and group them as a sub-caption $c_2$.

\begin{table*}[!t]
  \scriptsize
  \centering
  \caption{\textbf{Quantitative Comparison to SOTA Text-to-Image Models.}
  The best and second-best results are marked with \textcolor[HTML]{E38AAE}{Red} and \textcolor[HTML]{85BB65}{Green}.}
  \label{tab:quant_cmp}
  \begin{tabular}{>{\fontsize{10}{12}\selectfont}l
                    >{\fontsize{10}{12}\selectfont}c
                    >{\fontsize{10}{12}\selectfont}c
                    >{\fontsize{10}{12}\selectfont}c
                    >{\fontsize{10}{12}\selectfont}c
                    >{\fontsize{10}{12}\selectfont}c
                    >{\fontsize{10}{12}\selectfont}c
                    >{\fontsize{10}{12}\selectfont}c}
    \toprule
    \textbf{Methods} &  \textbf{FID$\downarrow$} & \textbf{HPSv2$\uparrow$} & \textbf{CLIP$\uparrow$}  & \textbf{Acc\textsubscript{obj}$\uparrow$} & 
    \textbf{Acc\textsubscript{tex}$\uparrow$} & \textbf{Acc\textsubscript{shape}$\uparrow$} & \textbf{Acc\textsubscript{all} $\uparrow$}  \\
    \midrule 
    SD 1.5~\cite{sd} & 48.09 & 0.2659 & \textcolor[HTML]{85BB65}{30.43} & 87.3 &77.4 &59.3 &74.6 \\
    SD 2.0~\cite{sd}   & 51.61 & 0.2588 & 26.27 &  82.8 & 74.7& 58.7 &72.0 \\
    SDXL~\cite{sdxl}   & 48.61 & 0.2647 & \textcolor[HTML]{E38AAE}{30.78} &  88.5  & 82.5 & 63.2 & 78.1 \\
    DeepFloyd-IF~\cite{deepfloyd}  & 44.62 & 0.2603 & 29.33   & 87.9 & 84.4 & 62.0 & 78.1 \\
    DALLE-2~\cite{dalle2}  & 49.60 & 0.2630 & 29.86  & 83.3 &  79.3 & 55.3 & 72.6 \\
    DALLE-3~\cite{dalle3}  & 66.36 & 0.2673 & 28.86  & 86.2 & 87.1 & 60.1 & 77.8 \\
    MidJourney~\cite{midjourney}  & 53.89 & 0.2688 & 28.89  & 85.2 & 79.5 & 59.4 & 74.7 \\
    \midrule
    \textbf{CosmicMan-SD}  & \textcolor[HTML]{85BB65}{36.78} & \textcolor[HTML]{85BB65}{0.2690}  & 28.47  
 & \textcolor[HTML]{85BB65}{91.7}  & \textcolor[HTML]{85BB65}{85.7} & \textcolor[HTML]{85BB65}{66.1} & \textcolor[HTML]{85BB65}{81.2}  \\
    \textbf{CosmicMan-SDXL}  & \textcolor[HTML]{E38AAE}{35.42}  & \textcolor[HTML]{E38AAE}{0.2698} & 27.31 &  \textcolor[HTML]{E38AAE}{92.7} & \textcolor[HTML]{E38AAE}{88.3} & \textcolor[HTML]{E38AAE}{69.7} & \textcolor[HTML]{E38AAE}{83.6} \\
    \bottomrule
  \end{tabular}
\end{table*}

\subsection{Decomposing and Refocusing Features} 
\label{hola}
During training diffusion models, the denoising loss $\mathcal{L}_{noise}$ can ensure the content generative capability of the model, but it lacks explicit alignment constraints between the caption and image pixels, especially when encountering descriptions with dense concepts that cover very high information density. Thus, on the shoulders of discrete human data mentioned in Sec.~\ref{hp}, we propose a new loss -- HOLA (short for {\textbf{H}uman Body and {\textbf{O}utfit} Guided {\textbf{L}oss} for {\textbf{A}lignment}) to seamlessly decompose the cross-attention features in SD model and enforce attention refocusing without adding extra modules.

Concretely, given the caption $C$ and latent $z_t$, the cross-attention maps $M$ can be decomposed as $M = (m_{(s_{1}, e_{1})}, m_{(s_{2}, e_{2})}, \ldots, m_{(s_{N}, e_{N})}, m_{other})$. Each $M_i$ is calculated through Eq.~\ref{attn}, with turning $K$ to $K_i$ (the projected embeddings of sub-caption $c_{(s_{i}, e_{i})}$).
We then incorporate HOLA alongside the original loss in SD to explicitly guide the cross-attention maps to have high responses only in specific regions. The HOLA is defined as follows:

\begin{scriptsize} 
\begin{equation}
 \mathrm{\mathcal{L}_{HOLA}} = \frac{1}{N} \sum_{i=1}^{N}(\sum_{j=s_{i}}^{e_{i}} \left \|  m_{j} - h_{i} \right \|_{2} ^{2} + \left \| \frac{1}{e_{i}-s_{i}} \sum_{j=s_{i}}^{e_{i}}(  m_{j}) - h_{i} \right \|_{2} ^{2}   )     
\end{equation}
\end{scriptsize}

Specifically, the first term of HOLA works under the guidance of human body structure -- it pushes the high response region of each concept feature to be as close as possible to the corresponding semantic region. However, since certain outfit-related concepts may only occupy a specific proportion within a semantic region, it is unnecessary to enforce their features to align with the whole semantic region. Also, concepts within the same group should be arranged harmoniously. Thus, we use the second term of HOLA to satisfy the situation. This term requires the average attention maps within one group to be close to their semantic map. It helps reduce ambiguities in outfit-level descriptions.
The overall loss function is as follows:
\begin{equation}
 \mathcal{L} = \alpha \mathcal{L}_{noise} + \beta \mathrm{\mathcal{L}_{HOLA}}
\end{equation}
where $\alpha$ and $\beta$ are hyper-parameters to balance the contribution of each loss.
\section{Experiments}
In this section, we compare our method with SOTA text-to-image (T2I) methods on human-centric generation tasks.
Sec.~\ref{sec:exper_set} shows experiment settings including implementation details and evaluation metrics.
Sec.~\ref{sec:cmp_sota} validates our method outperforms state-of-the-art T2I methods from quantitative and human preference evaluation.
Then, we provide an ablation study to evaluate the effectiveness of the design in training data and training strategy in Sec.~\ref{sec:abl}.
Finally, we show the pragmaticality and potential of CosmicMan as a foundation model in Sec.~\ref{sec:app}, by the application in two representative tasks in 2D and 3D respectively -- 2D human image editing and 3D human reconstruction.

\subsection{Experimental Settings} \label{sec:exper_set}
\noindent \textbf{Implementation Details.}
Our foundation models are based on Stable Diffusion (SD-1.5~\cite{sd} and SDXL~\cite{sdxl}).
We finetune the whole UNet from the pretrained model of Stable Diffusion combined with Daring framework on CosmicMan-HQ 1.0 dataset.
We use AdamW \cite{adamw} as the optimized method in 1e-5 learning rate and 1e-2 weight decay. Our model is trained on 32 80G NVIDIA A100 GPUs in a batch size of 64 for about one week.
Please refer to the supplementary material for more details.


\noindent \textbf{Evaluation Metrics.} \label{sec:metrics}
We evaluate results from three perspectives:
1) Image Quality: Frechet Inception Distance (FID)~\cite{fid} and Human Preference Score v2 (HPSv2)~\cite{hpsv2} are used to reflect diversity and authenticity.
2) Text-Image Alignment: CLIPScore~\cite{clipscore} provides a holistic measure of image-text alignment. 
However, it struggles to capture detailed image-text relationships, especially in fine-grained texture, shape, and object descriptions, which also be discussed in~\cite{dsg,compbench,geneval}.
Our proposed semantic accuracy metric, inspired by DSG~\cite{dsg}, enhances fine-grained text-image alignment, focusing on object (Acc\textsubscript{obj}), texture (Acc\textsubscript{tex}), shape (Acc\textsubscript{shape}), and overall (Acc\textsubscript{all}), making it suitable for human-centric evaluation.
3) Human Preference: we conduct a user study to evaluate the image quality and text-image alignment of each method.

\subsection{Comparison to Text-to-Image Models}
\label{sec:cmp_sota}
We compared our foundation model with various state-of-the-art text-to-image models, including open-source models like Stable Diffusion (SD-1.5/2.0), SDXL, DeepFloyd-IF, and commercial models such as DALLE2/3 and MidJourney. For a thorough comparison, we evaluated two versions of our foundation model: CosmicMan-SD based on SD-1.5 and CosmicMan-SXDL based on SDXL. 


\noindent \textbf{Quantitative Evaluation.}
We prepared a test set comprising $2048$ human images with fine-grained manually annotated prompts for fine-grained text-image generation.
We report the quantitative comparison in Tab.~\ref{tab:quant_cmp}. CosmicMan-SDXL excels in both image quality (FID) and fine-grained text-image alignment (Acc\textsubscript{all}). 
In terms of image generation quality, CosmicMan-SD/SDXL outperforms the corresponding SD-1.5/SDXL by a large margin, showing up to 23.52\% and 27.13\% relative improvements in FID.
As for fine-grained text-image alignment, CosmicMan-SDXL demonstrates superior generative capabilities in terms of three types of descriptions: object, texture, and shape. Compared to DALLE-3, which also unleashes the potential of detailed descriptions, CosmicMan shows 7.54\%, 1.38\%, 15.97\% and 7.46\% relative performance boost on Acc\textsubscript{obj}, Acc\textsubscript{tex}, Acc\textsubscript{shape} and Acc\textsubscript{all}.
Note that CosmicMan-SD/SDXL obtains a relatively low CLIPScore, as our emphasis was on evaluating fine-grained text-image alignment. In contrast, CLIPScore lacks the ability for fine-grained evaluation, consistent with the conclusions in the DSG~\cite{dsg} and GenEval~\cite{geneval}. CosmicMan-SD/SDXL achieves the best performance in Acc\textsubscript{all} and human preference evaluation, indicating its superiority in 2D human image generation.

\noindent \textbf{Human Preference Evaluation.}
We compared our results with DeepFloyd-IF, SDXL, DALLE3, and MidJourney through pairwise comparisons. 
The evaluation considered both image quality and text-image alignment, using 100 randomly selected prompts to generate corresponding images for each method. 
The evaluation results exhibits a significant preference, with over 93.06\%, 82.93\%, 78.13\% and 70.43\% of subjects favoring our results in terms of image quality, and 85.38\%, 90.25\%, 88.56\% and 81.68\% of subjects preferring our results over those of DeepFloyd-IF, SDXL, DALLE-3, and MidJourney in terms of text-image alignment. Qualitative results in the supplementary material further highlight our model's superiority in image quality, fine-grained details, and text-image alignment.

\subsection{Ablation Study} \label{sec:abl}

\noindent \textbf{Ablation on Training Data.} 
To show the validity of our proposed CosmicMan-HQ dataset, Tab.~\ref{tab:abl_data} reports the evaluation from three aspects: data source, data scale and annotation quality.
\noindent
1) Data Source. Compared with two cutting-edge datasets, LAION-5B and HumanSD, \textit{Ours} surpasses them by over 11.44 and 10.52 in FID, and 5.1 and 4.7 in Acc\textsubscript{all}, respectively. LAION-5B has large noise in both data and annotation, while HumanSD has fewer data quantities and coarse annotations. Owing to the scalable ability of our data production workflow, Annotate Anyone, the constructed CosmicMan-HQ dataset features a large quantity of high-quality annotations, which benefits the final results.
\noindent
2) Data Scaling. \textit{Ours}, trained with $6M$ images, brings a promotion of 2.51 in FID and 0.9 Acc\textsubscript{all} compared to $1M$ version \textit{Ours-3}, proving the effectiveness of data scaling. Thus, Annotate Anyone's capacity to run constantly to produce data is necessary to push the boundaries of foundation models' performance.
\noindent
3) Annotation Quality. We make a comparison under three different caption settings. \textit{Ours-3} with AA caption exhibits a significant improvement of 7.57 and 10.94 in FID, as well as 3.6 and 3.2 in Acc\textsubscript{all} compared to \textit{Ours-1} and \textit{Ours-2}. This verifies the effectiveness of improving the annotation quality of our proposed human-in-the-loop annotation mechanism in Annotate Anyone.

\begin{table}[!t]
  \tiny
  \centering
  \caption{\textbf{Ablation on Training Data.} ``AltText'' refers to Web Alternative Text, ``IB\textsubscript{pre}'' denotes the image descriptions generated by the pretrained InstructBLIP model, and ``Ours'' corresponds to captions produced by Annotate Anyone (``AA''). 
  }
  \label{tab:abl_data}
  \small
  \begin{tabular}{>{\fontsize{10}{12}\selectfont}l
                    >{\fontsize{10}{12}\selectfont}c
                    >{\fontsize{10}{12}\selectfont}c
                    >{\fontsize{10}{12}\selectfont}c
                    >{\fontsize{10}{12}\selectfont}c}
    \toprule
    \textbf{Dataset}  &  \textbf{Num}  &  \textbf{Text} & \textbf{FID$\downarrow$} &   \textbf{Acc\textsubscript{all} $\uparrow$} \\
    \midrule
    LAION-5B &  5\textit{B} & AltText & 48.09 &  74.6 \\
    HumanSD  &  1\textit{M} & AltText & 49.01 &  75.0 \\
    \midrule
    Ours-1 & 1\textit{M} & AltText & 47.65 &  75.2 \\
    Ours-2 & 1\textit{M} & IB\textsubscript{pre} & 51.02 &  75.6  \\
    Ours-3 & 1\textit{M} & AA &  40.08 & 78.8  \\
    \midrule
    \textbf{Ours} & 6\textit{M} & AA & \textcolor[HTML]{E38AAE}{37.57}  & \textcolor[HTML]{E38AAE}{79.7} \\
    \bottomrule
  \end{tabular}
\end{table}

\noindent \textbf{Ablation on Training Strategy.} Tab.~\ref{tab:abl_loss} shows the ablation of the training dataset and model design used in CosmicMan. By leveraging our CosmicMan-HQ dataset, fine-tuning the model gains a promotion of 10.52 in FID and 6.3 in Acc\textsubscript{all}. Our proposed $\mathrm{\mathcal{L}_{HOLA}}$ further enhances FID and Acc\textsubscript{all} by 0.79 and 1.5. Our novel perspectives on data and model design boost remarkable promotions of CosmicMan on fine-grained human generation.

\begin{table}[t]
  \footnotesize
  \centering
  \caption{\textbf{Ablation on Training Strategy.} ``Baseline'' refers to the SD pretrained model. ``CMHQ'' stands for CosmicMan-HQ.}
  \label{tab:abl_loss}
   \resizebox{1\linewidth}{!}{
  \begin{tabular}{>{\fontsize{10}{12}\selectfont}l
                    >{\fontsize{10}{12}\selectfont}c
                    >{\fontsize{10}{12}\selectfont}c
                    >{\fontsize{10}{12}\selectfont}c
                    >{\fontsize{10}{12}\selectfont}c
                    >{\fontsize{10}{12}\selectfont}c}
    \toprule
    \textbf{Methods}   & \textbf{FID$\downarrow$}  & \textbf{Acc\textsubscript{obj}$\uparrow$} &  \textbf{Acc\textsubscript{tex}$\uparrow$} & \textbf{Acc\textsubscript{shape}$\uparrow$} &  \textbf{Acc\textsubscript{all} $\uparrow$} \\
    \midrule
    Baseline  &  48.09 & 87.3  & 77.4 & 59.3 & 74.6 \\
    + CMHQ & 37.57 & 90.8  & 83.5 & 64.8 & 79.7 \\
    + {$\mathrm{\mathcal{L}_{HOLA}}$}  & \textcolor[HTML]{E38AAE}{36.78} & \textcolor[HTML]{E38AAE}{91.7}  & \textcolor[HTML]{E38AAE}{85.7} & \textcolor[HTML]{E38AAE}{66.1} & \textcolor[HTML]{E38AAE}{81.2} \\
    \bottomrule
   \end{tabular}
    }
\end{table}

\subsection{Applications}
\label{sec:app}
To validate the effectiveness of our human-specialized foundation model, we conduct additional experiments on 2D and 3D human-centric applications.

\noindent  \textbf{2D Human Editing.} 2D human editing manipulates human images for specified poses. We compare our CosmicMan-SDXL with SDXL based on T2I-Adapter~\cite{t2i-adapter}. In Tab.~\ref{tab:app}, our model outperforms SDXL on both FID and Acc\textsubscript{all}, 
showing its superiority in 2D human editing tasks.

\begin{table}[t] 
  \centering
  \caption{\textbf{Quantitative Comparison on 2D Human Editing and 3D Human Reconstruction.} User study reports the ratio of users who prefer our results to SD/SDXL.
  }
  \label{tab:app}
  \resizebox{1\linewidth}{!}{
  \begin{tabular}{lccc}
    \toprule
    \textbf{2D Application}     & \textbf{FID$\downarrow$}   &  \textbf{Acc\textsubscript{all} $\uparrow$} & \textbf{User Study}\\
    \midrule
    T2I-Adapter + SDXL  &  47.73  & 76.6 & 18.33\%\\
    \textbf{T2I-Adapter + CosmicMan-SDXL}  & \textcolor[HTML]{E38AAE}{37.62}   &  \textcolor[HTML]{E38AAE}{82.9} & \textcolor[HTML]{E38AAE}{81.67\%} \\
    \midrule
    \midrule
    \textbf{3D Application}        & \textbf{CLIP-Sim$\uparrow$}   &  \textbf{Acc\textsubscript{all} $\uparrow$} &  \textbf{User Study}\\
    \midrule
    Magic123 + SD & 0.83 & 67.6 & 26.36\%  \\
    \textbf{Magic123 + CosmicMan-SD} & \textcolor[HTML]{E38AAE}{0.88} & \textcolor[HTML]{E38AAE}{70.8}  &  \textcolor[HTML]{E38AAE}{73.64\%} \\
    \bottomrule
  \end{tabular}
  }
\end{table}

\noindent \textbf{3D Human Reconstruction.} 
We validate the effectiveness of our CosmicMan-SD model based on Magic123~\cite{magic123}, one representative 3D object reconstruction method from a single image.
We replace the SD pretrained model with our foundation model in Magic123 for comparison. 
The higher CLIP-similarity~\cite{magic123} and Acc\textsubscript{all} in Tab.~\ref{tab:app} exhibit the superior potential of our model on 3D human reconstruction.

\section{Discussion}
\noindent
\textbf{Release.}
We seriously treat the license and privacy issues and follow a rigorous legal review in our institute. CosmicMan-HQ 1.0 with all of the annotations will be released step by step. People in the dataset are anonymized without additional private or sensitive metadata. All released data are free for research use only. The model and codes will also be released.

\noindent
\textbf{Future Work.}
Not placing CosmicMan merely as a research paper, we also commit ourselves to providing a long-term and sustainable foundation platform to support the research in human-centric content generation. Thus, we will continuously 1) operate Annotate Anyone to produce subsequent versions of CosmicMan-HQ aligned dynamically with real-world data, and 2) provide up-to-date human-specialized foundation models periodically trained on new versions of our data. By providing a well-constructed and long-term-maintained infrastructure, we hope to benefit broader research communities centered on human subjects.

{
    \small
    \bibliographystyle{ieeenat_fullname}
    \bibliography{main}
}

\clearpage
\appendix
\setcounter{figure}{0}
\noindent
\textbf{\LARGE Appendix}
\vspace{5ex}

This appendix serves to further enrich the discourse established in our main paper. In Sec.~\ref{aa}, we present additional elaboration on the scalable data production pipeline -- Annotate Anyone. Then, in Sec.~\ref{method}, we delve into the proposed training framework (\textit{i.e.,} Daring) with additional evaluation and illustrations. Lastly, we present a broader range of qualitative results, including comparisons with state-of-the-art methods, more examples of generated samples, extensive ablation studies, and potential application demonstrations in Sec.~\ref{exp}. 


\section{Annotate Anyone}\label{aa}
As introduced in the main paper, the whole data production paradigm comprises two parts that benefit from the cooperation of human effort and AI models -- a flowing data sourcing for data collection, and human-in-the-loop data annotation. In this section, we unfold the details of Annotate Anyone. We first complement the data filtering step in Sec.~\ref{sec:datafiltering_supp}. Then, we provide the pseudo-code for fine-tuning the VLM model used in Annotate Anyone, accompanied by an exhaustive explanation (Sec.~\ref{sec:pseudo}). Besides, we conduct experiments to demonstrate the effectiveness of the proposed human-in-the-loop annotation pipeline in Sec.~\ref{sec:ib_ablation_supp}. Moreover, we provide a detailed exposition of all $70$ questions we used on VLM for obtaining the description of an image, as presented in Sec.~\ref{sec:label_prot_supp}. We also list the corresponding attribute labels extracted from the output of VLM models. 
Finally, a detailed statistical analysis (Sec.~\ref{sta}) and additional data samples from our dataset CosmicMan-HQ 1.0 (Sec.~\ref{mds}) are presented.

\begin{algorithm}[b]
\caption{Annotate Anyone: \\Iteratively improve VLM model to annotate image}
\label{arg:annotate_anyone}
\begin{algorithmic}
    \State \textbf{Input: } $I$ \Comment{Images in Data Pool}
    \State $I_0 \gets \text{sample test set from } I$ \Comment{K images}
    \State $A_0 \gets \text{label } I_0 \text{ by VQA pretrained model (IB}_0\text{)}$
    \State $M_0 \gets \text{label } I_0 \text{ by human}$
    \State $\text{Acc} \gets \text{Eval}(A_0, M_0)$ \Comment{Acc for all label}
    \State $i=0$
    \State \textbf{while }$\text{Acc} < 85\%$ do 
        \State \hspace{\algorithmicindent} $i \gets i+1$
        \State \hspace{\algorithmicindent} $I_i \gets \text{sample K images from data pool}$
        \State \hspace{\algorithmicindent} $\tilde{M}_i \gets \text{select labels from } M_i\text{with Acc } < 85\%$
        \State \hspace{\algorithmicindent} $\text{IB}_{i+1} \gets \text{finetune IB}_i \text{with (} \cup_{1}^{i} \tilde{M}_i, I_i \text{)}$
        
        \State \hspace{\algorithmicindent} $A_{i+1} \gets \text{label } I_0 \text{ by (IB}_{i+1}\text{)}$

        \State \hspace{\algorithmicindent} $\text{Acc} \gets \text{Eval}(A_{i+1}, M_0)$ 
    \State \textbf{end while} 
    \State  $\text{(}I, A\text{)} \gets \text{update all image label in data pool using } \text{IB}_\text{i+1}$
\end{algorithmic}
\end{algorithm}

\subsection{Data Filtering}
\label{sec:datafiltering_supp}
As mentioned in the main text, we use three academic datasets, LAION-5B~\cite{laion5b}, SHHQ~\cite{sghuman}, and DeepFashion~\cite{deepfashion} as part of data origin. When distilling LAION-5B dataset, we extensively utilize the meta information provided by LAION-5B itself to perform an initial screening of the data. We filter out a significant amount of images that contain the ``Not safe for work (nsfw)'' label or have a watermark score higher than 0.5. The data source of SHHQ and DeepFashion is relatively concentrated, consisting exclusively of watermark-free fashion / studio-shot data. Thus, we omit this step for those two datasets.
Before fetching images from the Internet, we excluded the data origins of LAION-5B, SHHQ, and DeepFashion. 
Our aim is to minimize crawling data from the same source as much as possible.
We also remove duplicated images by encoding images into fixed-length hash values using perceptual hashing algorithms~\cite{imagededup}. Images with hamming distances smaller than 3 are considered duplicates, and one of them will be removed to avoid redundancy.


\subsection{Human-in-the-loop Data Annotation}
\label{sec:pseudo}
This section presents the pseudo-code of the annotation algorithm (see Algorithm.~\ref{arg:annotate_anyone}) mentioned in Sec.~\ref{sec:human-in-the-loop} of the main text for Annotate Anyone. A detailed step-by-step explanation will be provided below.

\begin{figure*}[t]
    \centering
    \includegraphics[width=1\textwidth]{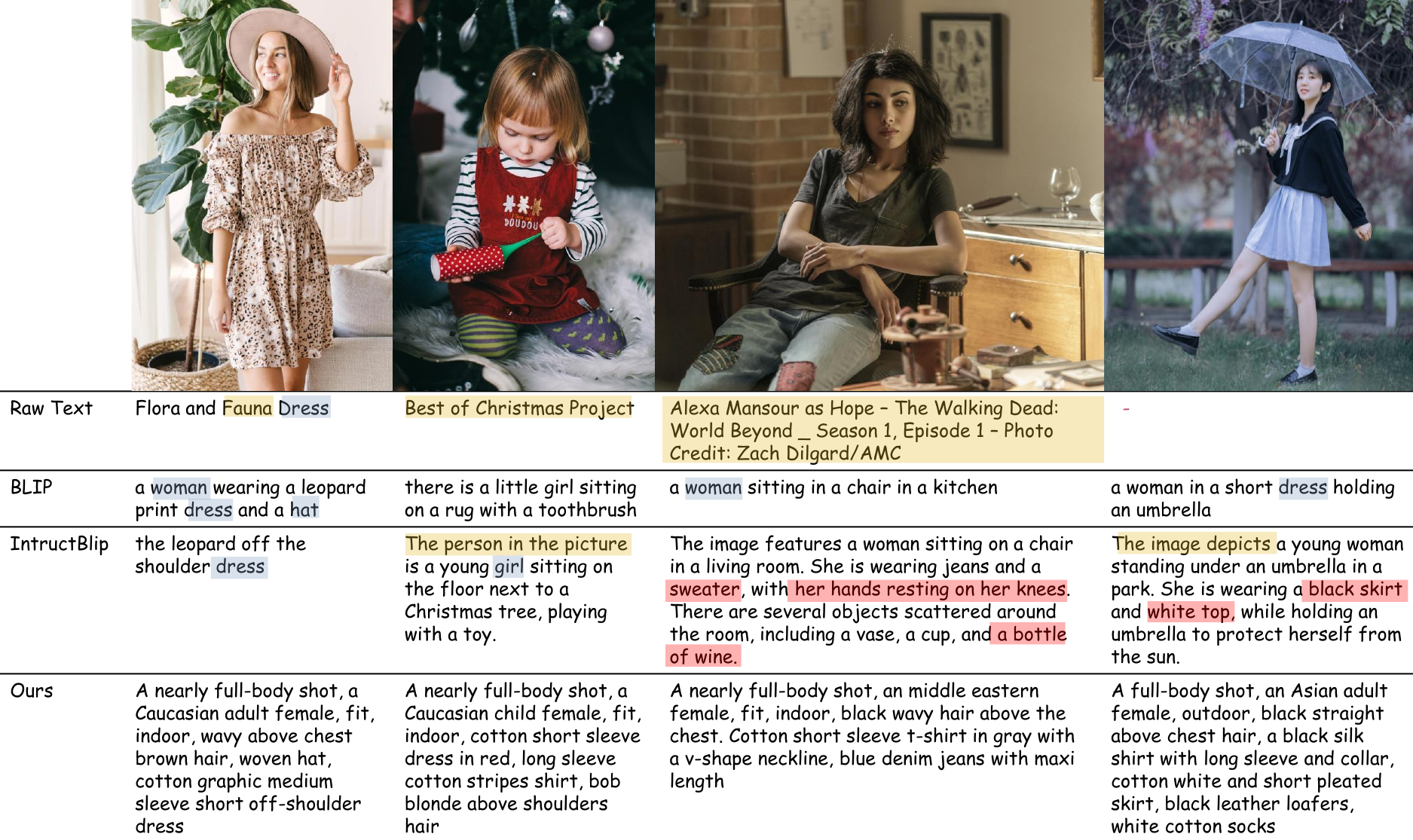}
    \caption{\textbf{Data and Annotation Samplings.} Sample images with captions obtained from different methods: the original text uploaded along with the images, captions generated by BLIP~\cite{blip}, and InstructBLIP~\cite{instructblip} pretrained models, and our attribute-based text. Here we use different colors to highlight the \colorbox{highlightyellow}{\textcolor{black}{unrelated}}, \colorbox{highlightred}{\textcolor{black}{wrong}}, and \colorbox{highlightblue}{\textcolor{black}{coarse and vague}} descriptions.
    }
    \label{fig:data_and_annotation_samplings}
\end{figure*}

\begin{figure}[t]
    \centering
    \includegraphics[width=1\linewidth]{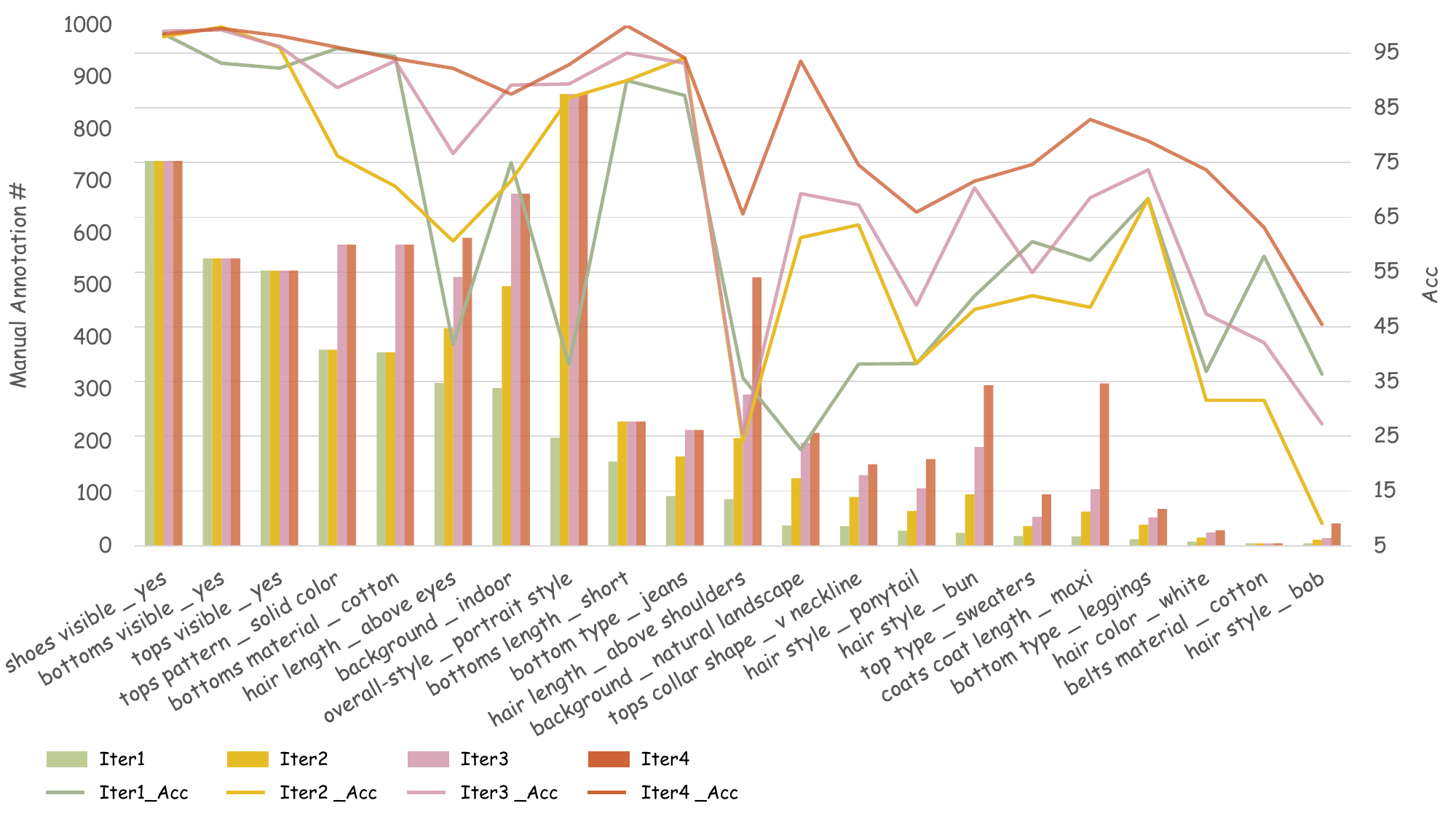}
    \caption{\textbf{Effectiveness of Annotate Anyone.} The bar charts represent the manual annotation counts for different attributes in $1000$ images for each iteration, and the solid lines depict the accuracy of each attribute under the current model. Better zoom in for details. 
    }
    \vspace{-0.5cm}
    \label{fig:ib_ablation2}
\end{figure}

The overall objective of this process is to achieve collaboration between AI and human to provide descriptions for all images (denoted as $I$) within the entire data pool. 
We start the workflow by randomly selecting a fixed test set $I_0$, consisting of $K$ images from the data pool $I$. $I_0$ will be asked $70$ meticulously crafted vision questions (see Tab.~\ref{tab:detailed_questions} for reference) by the pretrained model $IB_0$~\cite{instructblip} and obtain corresponding answers $A_0$.
Note that, inspired by the annotation approach used in Fashion datasets~\cite{deepfashion,fashionpedia}, we transform the uncontrollable, attributes highly entangled captioning tasks into detailed question-answering tasks to get $A_0$. This approach results in structured label answers, proceeding from coarse to finer granularity, that can ease evaluation and provide more accurate and fine-grained region-specific labels. 
Then, we ask a professional annotation team to manually label the images in $I_0$ based on these $70$ questions, to form the attribute dictionary $M_0$ as the ground-truth of the test set.
For every single question, we evaluate the accuracy of the pretrained model $IB_0$ by comparing the label in $A_0$ and $M_0$.
\begin{figure*}[t]
    \centering
    \includegraphics[width=1\linewidth]{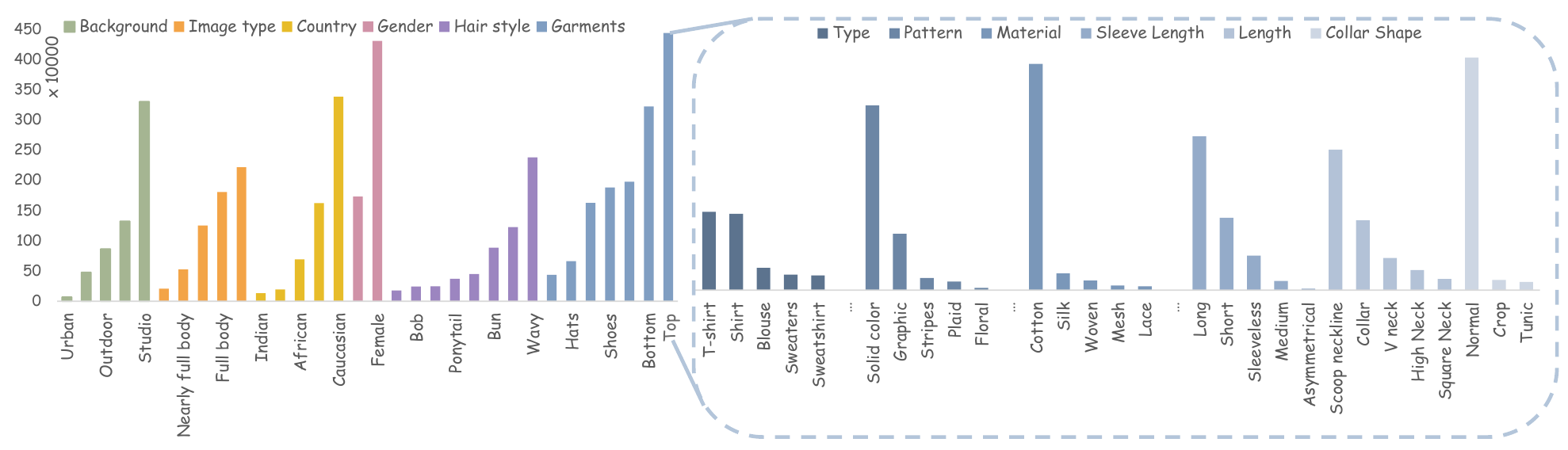}
    \vspace{-0.5cm}
    \caption{\textbf{CosmicMan-HQ 1.0 Statistics.} The left part shows the global attributes and garment types of the person in the image, and the right section displays attributes examples of top garments. }
    \label{fig:data_stats}
\end{figure*}

\begin{figure*}[t]
    \centering
    \includegraphics[width=1\linewidth]{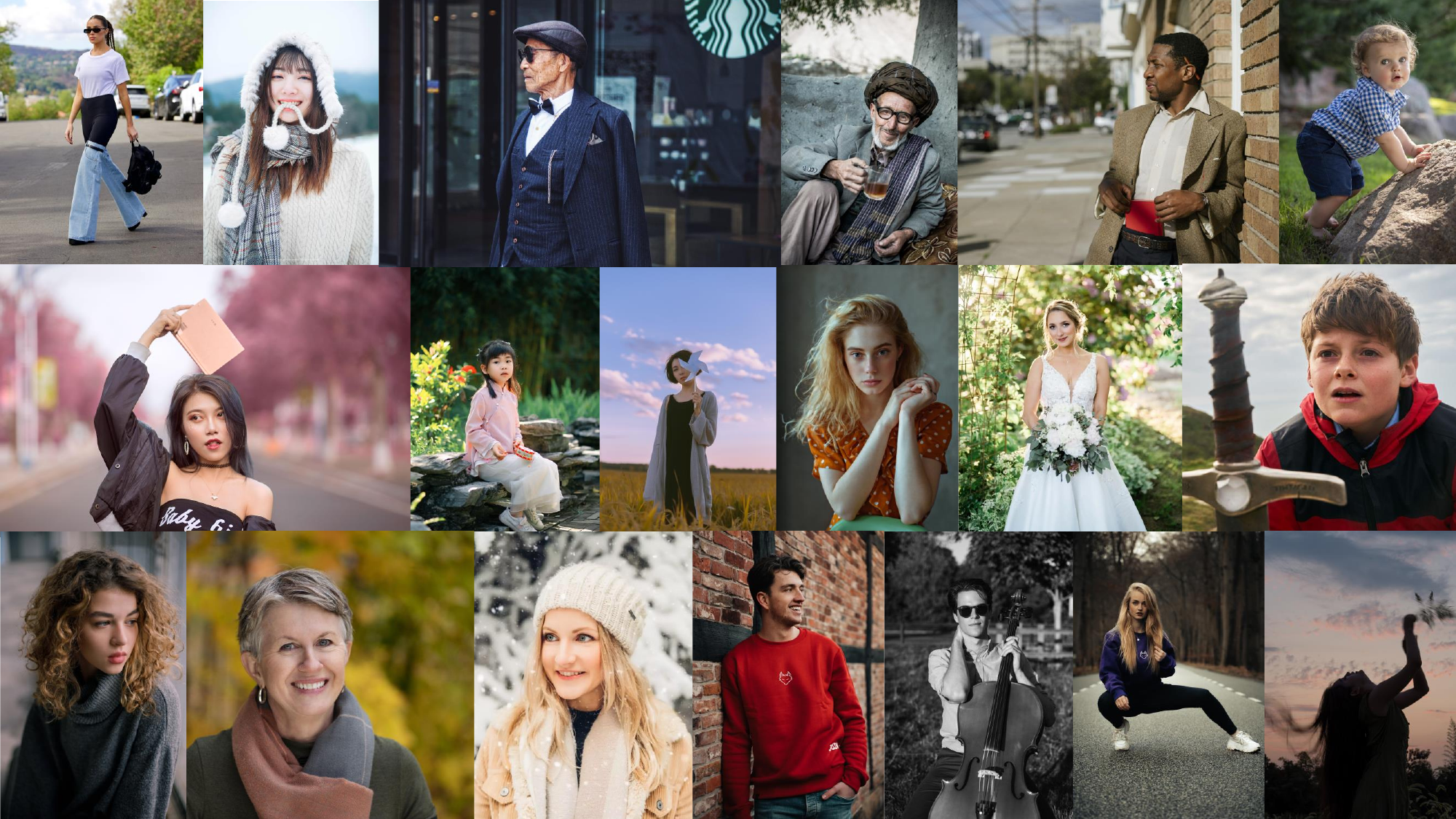}
    \caption{\textbf{CosmicMan-HQ 1.0 Image Examples.} The examples display the dataset's wide-ranging diversity, encompassing different aspects like age, ethnicity, and clothing styles.}
    \vspace{-0.5cm}
    \label{fig:moresamples}
\end{figure*}

The label accuracy for every attribute corresponding to every question will be documented and utilized to assess whether the current (pretrained and fine-tuned) model's judgment for that attribute meets the criteria. If there is one label's accuracy less than $85\%$, the fine-tuning loop of AI model $IB$ starts. Concretely, We will randomly sample another $K$ images from the data pool, and ask the annotation team to label the attributes whose model accuracy is less than $85\%$.
After each round of manual annotation, the newly manual-labeled data is combined with the previously annotated data and utilized for fine-tuning the $IB_i$ model.
During training, we use `` \textless Image \textgreater: \{\}. Question: \{\} Answer:" as a template to construct fine-tuning dataset. The fine-tuned model $IB_{i+1}$ will be utilized to perform inference on the test set $I_0$ for evaluation to get $Acc$.
This fine-tuning process will repeat until the overall label accuracy achieves $85\%$. The final model $IB_{i+1}$ is used to annotate all the images in the data pool, and results image-label pairs ($I$, $A$).

\begin{table*}[tp]
\footnotesize
\caption{\textbf{The Detailed Questions Asked to Describe Every Single Image.} The index 0-2 in Acc represents the corresponding questions is evaluated as \textbf{Acc\textsubscript{obj}, \textbf{Acc\textsubscript{tex}}, and \textbf{Acc\textsubscript{shape}}. }}
\label{tab:detailed_questions}
\centering
\resizebox{0.85 \linewidth}{!}{
\begin{tabular}{l|c|c|l|l}
    \toprule
    & Q\textsubscript{idx} & Acc & Questions & Attributes \\ \hline
    \multirow{6}{*}{Overall} & 1 & 0 & what is the gender of the person in the image?  & male, female \\
    &2&0& what is the country of the person in the image? & Indian, Latino, Middle Eastern, ... \\
    &3&0& what is the age of the person in the image? & teenager, adult, elderly, ...  \\
    &4&0& what is the body shape of the person in the image? & fit, skinny, obese, muscular \\
    &5&0& what is the background of the person in the image? & \\
    &6&0& what is the overall-style of the person in the image? & fashion, documentary, portrait, others \\ \hline
    \multirow{4}{*}{Hair} & 7 & 0 & is the hair of the person visible in the image? & yes, no \\
    &8&-& what is the hair color of the person? & brown, gray, violet, ...\\ 
    &9&1&what is the hairstyle of the person? & wavy, ponytail, straight, ... \\
    &10&2&what is the hair length of the person? & below chest, bald, bob, ... \\
    \hline
    \multirow{7}{*}{\centering\parbox{2cm}{Top\\Clothings}} &11&0& does the person wears any tops? & yes, no \\ 
    &12&0&what kind of top does the person wear? &vest, blouse, hoodie, ... \\
    &13&1&what is the pattern of the tops? & graphic, printed, stripes, ...\\
    &14&1&what is the material of the tops? &linen, fur, lace, ... \\
    &15&2&what is the sleeve length of the tops? & short, medium, sleeveless, ... \\
    &16&2&what is the top length of the tops? & crop, normal, tunic\\
    &17&2&what is the collar shape of the tops?& square, v-shape, collar, ... \\  
    &18&-&are the top clothing \{random color\}? & yes, no \\ \hline
    
    \multirow{6}{*}{\parbox{2cm}{Bottom \\ Clothings}} &19&0& does the person wears any bottoms? &  yes, no \\ 
    &20&0&what kind of bottom does the person wear? &shorts, sweatpants, skirt, ... \\
    &21&1&what is the pattern of the bottoms? &graphic, printed, stripes, ... \\
    &22&1&what is the material of the bottoms? & linen, fur, lace, ... \\
    &23&2&what is the length of the bottoms? &  short, medium, sleeveless, ... \\
    &24&2&what is the bottom shape of the bottoms? & straight, tapered, wide-leg, ...\\  %
    &25&-&are the bottom clothing \{random color\}? & yes, no \\ \hline
    
    \multirow{8}{*}{\centering\parbox{2cm}{One-piece\\Outfits}} &26&0& does the person wears any one-piece outfits?& yes, no\\ 
    &27&0&what kind of one-piece outfit does the person wear? &bathrobe, jumpsuit, dress, ...\\
    &28&1&what is the pattern of the one-piece outfits?  & graphic, printed, stripes, ... \\
    &29&1&what is the material of the one-piece outfits? & linen, fur, lace, ... \\
    &30&2&what is the sleeve length of the one-piece outfits? & short, medium, sleeveless, ... \\
    &31&2&what is the collar shape of the one-piece outfits? & square, v-shape, collar, ... \\
    &32&2&what is the length of the one-piece outfits? &  short, medium, sleeveless, ... \\
    &33&2&what is the shoulder exposure level of the one-piece outfits?  &one-shoulder, off-shoulder \\ \hline
    
    \multirow{6}{*}{Coats}&34&0&does the person wears any coats?&yes, no\\ 
    &35&0&what kind of coat does the person wear?&cape, trench coat, blazer, ... \\
    &36&1&what is the pattern of the coats? & graphic, printed, stripes, ... \\
    &37&1&what is the material of the coats?  & linen, fur, lace, ... \\
    &38&2& what is the coat length of the coats?&short, medium, maxi\\
    &39&2& what is the collar shape of the coats? &  square, v-shape, collar, ... \\ 
    &40&-& are the coat \{random color\}? & yes, no \\ \hline
    \multirow{3}{*}{\centering\parbox{2cm}{Special\\Clothings}} &41&0&does the person wears any special clothings? & yes, no\\ 
    &42&0& what kind of special clothing does the person wear?& Hanfu, Saree, cosplay, ...\\  
    &43&2& what is the sleeve length of the special clothing?& short, medium, sleeveless, ... \\\hline
    
    \multirow{6}{*}{Shoes} &44&0&does the person wears any shoes? & yes , no\\ 
    &45&0&what kind of shoe does the person wear?&sneakers, flip flops, loafers, ...\\
    &46&1&what is the pattern of the shoes? & graphic, printed, stripes, ... \\
    &47&1&what is the material of the shoes? & linen, fur, lace, ... \\
    &48&2&what is the boots length of the shoes?&ankle, mid-calf, knee-high, ...\\ 
    &49&-& are the shoes \{random color\}? & yes, no \\ \hline
    
    \multirow{3}{*}{Bags} &50&0&does the person wears any bags? &yes, no \\
    &51&0& what is the type of the bags?& tote bag, handbag, wallet, ...\\
    &52&1& what is the material of the bags?& linen, fur, lace, ...\\ \hline
    \multirow{3}{*}{Hats} &53&0&does the person wears any hats? & yes, no \\
    &54&0& what is the type of the hats?& beret, sun hat, helmet, ...,\\
    &55&1& what is the material of the hats? & linen, fur, lace, ...\\ \hline
    \multirow{2}{*}{Socks}&56&1& what is the pattern of the socks? & graphic, printed, stripes, ...\\
    \multirow{6}{*}{\centering\parbox{2cm}{Other\\Accessories}} &57&1& what is the material of the socks? & linen, fur, lace, ... \\ \hline
    &58&1& what is the pattern of the belts? & graphic, printed, stripes, ... \\
    &59&1& what is the material of the scarf?& linen, fur, lace, ... \\
    &60&1& what is the pattern of the scarf? & graphic, printed, stripes, ... \\ 
    &61&1& what is the material of the ties?& linen, fur, lace, ...\\
    &62&1& what is the pattern of the ties? & graphic, printed, stripes, ...\\\hline
    \multirow{8}{*}{Headwear}&63&0& does the person wears any headwear? & yes, no\\
    &64&0& what kind of headwear does the person wear? & headband, headscarf, veil \\
    &65&1& what is the pattern of the headband? & graphic, printed, stripes, ...\\
    &66&1& what is the material of the headband? & linen, fur, lace, ...\\
    &67&1& what is the pattern of the headscarf? & graphic, printed, stripes, ...\\
    &68&1& what is the material of the headscarf? & linen, fur, lace, ...\\
    &69&1& what is the pattern of the veil? & graphic, printed, stripes, ...\\
    &70&1& what is the material of the veil? &  linen, fur, lace, ...\\
    \bottomrule
\end{tabular}
}
\end{table*}

\subsection{Effectiveness of Annotate Anyone}
\label{sec:ib_ablation_supp}
In this section, we conduct experiments to assess the effectiveness of reducing manual annotation efforts in our Annotate Anyone pipeline.

We run four sets of model fine-tuning with a step-wise augmentation of manually labeled data. In every iteration, we randomly sampled $1000$ images ($K$ in Algorithm.~\ref{arg:annotate_anyone} is set to $1000$) from $I$ in the data pool. Leveraging evaluation metrics (Acc) from the previous model, we progressively decrease the need for manual labeling by excluding already qualified attributes (where the Acc is higher than $85\%$). The annotated data accumulates and is utilized for fine-tuning the subsequent model.

As shown in bar charts of the first iteration in Fig.~\ref{fig:ib_ablation2}, due to the random sampling from the data pool, the inherent attribute distribution of sampled images naturally exhibits a long-tail pattern similar to that of CosmicMan-HQ 1.0 (Fig.~\ref{fig:data_stats}). The accuracy of different attributes also correlates to the number of training data. Then in the following iterations, as we only augment the attributes that did not meet the accuracy threshold, the long-tail distribution of the training data is alleviated, as shown in the bar charts of Fig.~\ref{fig:ib_ablation2}. 
Specifically, with the progression of iterations, the number of labels requiring manual annotation on each image gradually decreases. Results in the figure showcase that the tail-end labels have improved over iterations while the existing qualified labels maintain their accuracy. The overall accuracy of the final fine-tuned model has significantly increased from $56.0\%$ in the pre-training phase to $85.3\%$. We have temporarily halted this flywheel because the average accuracy has reached $85\%$, but it can continue run until the accuracy of every single attribute meets the standard.

\subsection{Label Protocol}
\label{sec:label_prot_supp}
In Tab.~\ref{tab:detailed_questions}, we present a series of questions related to each part from the human parsing map, resulting in a total of $70$ question-answer pairs to describe a human image.

\subsection{Statistics of CosmicMan-HQ 1.0}
\label{sta}
To visually demonstrate the superiority of our dataset in terms of caption granularity and accuracy, 
Fig.~\ref{fig:data_and_annotation_samplings} presents images from CosmicMan-HQ 1.0 paired with annotations from different sources. As shown, the text, including the raw caption from original websites and generated caption from the BLIP~\cite{blip}, often lacks descriptive details. Raw text is often unrelated to the image content, while text generated by BLIP is correct in general but vague,
and the captions from the InstructBLIP~\cite{instructblip} have lower accuracy in providing detailed captions, particularly in attribute-level descriptions. For example, InstructBLIP incorrectly stated that the woman is wearing a sweater instead of a short-sleeved shirt, as shown in the description of the third image in Fig.~\ref{fig:data_and_annotation_samplings}.
Compared to these captions, our method provides comprehensive and accurate descriptions by transforming fine-grained labels using the fine-tuned IB model.

Following that, we delve into an in-depth exploration of the data distribution within the dataset. The statistical analysis of the overall image and human-centric attributes across CosmicMan-HQ 1.0 is displayed in the left half of Fig.~\ref{fig:data_stats}. The right half offers examples of fine-grained attributes related to ``top clothing'', which further demonstrates the granularity of our annotation. The histogram presents a wide coverage of various attributes and a natural long-tail distribution, which reflects our dataset's consistency with the data in the real world. 

\subsection{More Dataset Samples} 
\label{mds}
In Fig.~\ref{fig:moresamples}, more image examples are randomly sampled from CosmicMan-HQ 1.0, which showcases the dataset's inherent diversity across various dimensions, such as age, ethnicity, and garments. 

\section{Daring}\label{method}
In this section, we primarily focus on the additional analysis of our proposed training framework, \textit{Daring}.
Regarding the Data Discretization, we supplement with an ablation study on the type of captions, validating the effectiveness of the decomposed text space in Sec.~\ref{sec:data_discretization}. We then provide a comparison on the two terms of the proposed HOLA loss in Sec.~\ref{sec:loss_hola}. 
By presenting the cross-attention maps of the proposed model, we elucidate that our method improves the text-image alignment for dense concepts in Sec.~\ref{sec:attn_map}.
We also conduct an experimental evaluation of optimization strategies involving different loss functions in Sec.~\ref{exp:mmo}.

\subsection{Evaluation of Data Discretization}
\label{sec:data_discretization}

Our framework first decomposes both the text and image into several groups of pair data. The underlying motivation is that we hypothesize the descriptions are decomposable and finite for describing human images, since they are inherently associated with human body structure, and each concept can correspond to a particular body region.  To evaluate the effectiveness of Data Discretization, we conduct a experiment on decomposed and continuous text space  as shown in Tab.~\ref{tab:textspace}. 
Of note, to cost-effective get natural continuous text, we use GPT-4 to generate 5-10 templates for each set of labels in the human structure, such as: ``\{type\} with \{pattern\} made of \{material\}, \{sleeve\_length\} sleeves, \{collar\_shape\} neckline, \{length\} length, \{shoulder\_exposure\_level\} exposure'' for ``One-piece Outfits'' group. 
The natural descriptions of each group are concatenated to get a sentence for training and testing.
Since the text encoder of SD can only handle 77 tokens, we construct another test set with short captions.
For each sample within the original test set, when the number of tokens within the natural description is greater than 77, we randomly dropout descriptions of certain body parts.
It can be seen that with the decomposed text data, the \textbf{Acc\textsubscript{all}} exhibits a significant improvement.
This is because decomposed text space ensures that all important details are accurately and clearly included and does not introduce ambiguity or unnecessary context.
While data discretization gives an additional boost in performance, the best performance is achieved when $\mathrm{\mathcal{L}_{HOLA}}$ and data discretization are combined. 
This illustrates that as long as keys $K$ in attention blocks are initially decomposable and semantically guided, it may not be necessary to optimize latent code using cross-attention maps in the inference or add complex modules to SD's original architecture for dense concepts alignment.

\begin{table}[H]
  \setlength{\tabcolsep}{1pt}
  \footnotesize
  \centering
  \caption{\textbf{Ablation on $\mathrm{\mathcal{L}_{HOLA}}$ and Data Discretization.} The best results are marked with \textcolor[HTML]{E38AAE}{Red}.}
    \begin{tabular}{>{\fontsize{10}{12}\selectfont}c
                    >{\fontsize{10}{12}\selectfont}c
                    >{\fontsize{10}{12}\selectfont}c
                    >{\fontsize{10}{12}\selectfont}c
                    >{\fontsize{10}{12}\selectfont}c
                    >{\fontsize{10}{12}\selectfont}c}
    \toprule 
        $\mathrm{\mathcal{L}_{HOLA}}$ & Discretization  &
        \textbf{Acc\textsubscript{obj}$\uparrow$} &  \textbf{Acc\textsubscript{tex}$\uparrow$} & \textbf{Acc\textsubscript{shape}$\uparrow$} &  \textbf{Acc\textsubscript{all} $\uparrow$} \\
    \midrule
         & & 74.6 & 88.4 &56.2  & 73.1 \\
         \cmark & & 76.8 & 89.3 & 59.3 & 75.1 \\
         & \cmark  & 82.4 & 91.3 & 66.5 & 80.1 \\
        \cmark  & \cmark & \textcolor[HTML]{E38AAE}{85.5} & \textcolor[HTML]{E38AAE}{92.2} & \textcolor[HTML]{E38AAE}{71.4} & \textcolor[HTML]{E38AAE}{83.1} \\
    \bottomrule
    \end{tabular}
  \label{tab:textspace}
\end{table}

\subsection{Evaluation of HOLA Loss}
\label{sec:loss_hola}

The proposed loss $\mathrm{\mathcal{L}_{HOLA}}$ for guiding the model to align with the region-level captions is composed of two terms. The first term introduces the overall spatial structure and single concept, while the second term reduces the ambiguities of outfit-level descriptions. For example, patterns exhibited on clothing typically manifest in specific areas rather than uniformly across the entire garment. Consequently, these patterns do not necessitate alignment across the entire clothing piece. Here we also provide quantitative results in Tab.~\ref{tab:loss-term}. 
Baseline means the model that only fine-tuned on CosmicMan-HQ-1M.
The employment of ``term1'' markedly enhances all semantic accuracy. Subsequent incorporation of ``term2'' yielded a more pronounced enhancement in \textbf{Acc\textsubscript{tex}} due to the fact that certain outfit-level texture attributes such as patterns don't need to be aligned with the whole semantic region.

\begin{table}[H]
  \centering
  \caption{\textbf{Ablation on Loss Terms in $\mathrm{\mathcal{L}_{HOLA}}$. } Term1 pushes the high response region of each single concept feature. Term2  helps reduce the ambiguities of outfit-level descriptions.}
  \begin{tabular}{>{\fontsize{10}{12}\selectfont}l
                    >{\fontsize{10}{12}\selectfont}c
                    >{\fontsize{10}{12}\selectfont}c
                    >{\fontsize{10}{12}\selectfont}c
                    >{\fontsize{10}{12}\selectfont}c}
    \toprule
    \textbf{Methods}   &  \textbf{Acc\textsubscript{obj}$\uparrow$} &  \textbf{Acc\textsubscript{tex}$\uparrow$} & \textbf{Acc\textsubscript{shape}$\uparrow$} &  \textbf{Acc\textsubscript{all} $\uparrow$} \\
    \midrule
    Baseline   & 87.3 & 77.4 & 59.3  & 74.6 \\
    + term1   & 91.3 & 84.7 & 65.6  & 80.5 \\
    + term2   & \textcolor[HTML]{E38AAE}{91.7}  & \textcolor[HTML]{E38AAE}{85.7} & \textcolor[HTML]{E38AAE}{66.1} & \textcolor[HTML]{E38AAE}{81.2} \\
    \bottomrule
  \end{tabular}
  \label{tab:loss-term}
\end{table}

\begin{figure*}[!t]
    \centering
    \includegraphics[width=0.95\textwidth]{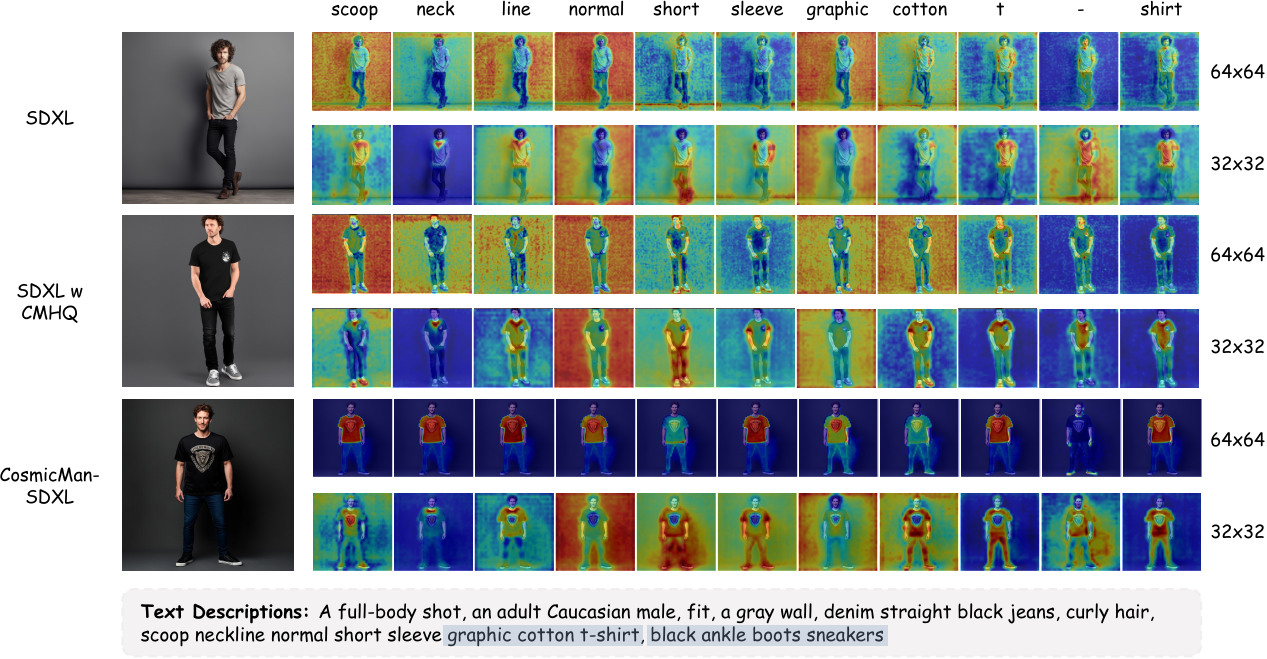}
    \caption{\textbf{Visualization of Cross-attention Maps.} We show cross-attention maps in SDXL-base for two different resolutions (map of 64$\times$64 shape with 16$\times$ down-sampling and map of 32$\times$32 shape with 32$\times$ down-sampling), in a similar way to the Prompt-to-Prompt \cite{p2p} visualization method. For a clearer comparison, we further resize them to the original image size.}
    \label{fig:attn}
\end{figure*}

\subsection{Evaluation of Cross-Attention Maps}
\label{sec:attn_map}
We also provide the comparison of the cross-attention maps across various models, as illustrated in Fig.~\ref{fig:attn}.
In summary, our evaluation indicates that CosmicMan-SDXL exhibits enhanced precision in the generation of graphic t-shirt and black sneakers. This is further corroborated by our detailed examination of the attention maps for ``Top Clothing''. Notably, the attention map corresponding to the ``graphic'' attribute demonstrates a more accurate activation in the top region, particularly when compared with the attention map extracted from SDXL. Furthermore, a comprehensive comparison of all attributes shows that CosmicMan-SDXL not only achieves higher semantic accuracy, but also achieves more accurate semantic activation in the corresponding attention map than the model fine-tuned on CosmicMan-HQ.

\subsection{Evaluation of Optimization Strategies}
\label{exp:mmo}

In addition to appending a loss term through linearization and adjusting scalar coefficients, we conduct additional experiments employing a multi-objective optimization technique, Random Loss Weighting~(RLW)~\cite{RLW}, on CosmicMan-SD. The results presented in Tab.~\ref{tab:multi_object} reveal that both using RLW directly and combining it with a weighted loss term lead to a decline in model performance. 

\begin{table}[H]
    \centering
    \caption{\textbf{Ablation on Optimization Strategies.} 
    The best results are marked with \textcolor[HTML]{E38AAE}{Red}.}
    \renewcommand{\arraystretch}{1}        \resizebox{1\linewidth}{!}{

    \begin{tabular}{ccccccc}
    \toprule
      RLW  &  $\beta$ & FID$\downarrow$ & \textbf{Acc\textsubscript{obj}$\uparrow$} &  \textbf{Acc\textsubscript{tex}$\uparrow$} & \textbf{Acc\textsubscript{shape}$\uparrow$} &  \textbf{Acc\textsubscript{all}$\uparrow$ } \\
    \midrule
     \checkmark &  & 55.5 & 83.7 & 76.7 & 60.8 & 73.6 \\
     \checkmark & \checkmark & 39.9 & 89.9 & 81.7 & 63.2 & 78.2 \\
       & \checkmark  & \textcolor[HTML]{E38AAE}{36.8} & \textcolor[HTML]{E38AAE}{91.7} & \textcolor[HTML]{E38AAE}{85.7} & \textcolor[HTML]{E38AAE}{66.1} & \textcolor[HTML]{E38AAE}{81.2}   \\
    \bottomrule
    \end{tabular}
    }
    \label{tab:multi_object}
\end{table}

\section{Experiments}\label{exp}
To more vividly demonstrate the performance of our text-to-image foundation model, CosmicMan, this section provides an extensive array of qualitative results. In Sec.~\ref{exp:setting} and Sec.~\ref{exp:metric}, we begin by detailing our experimental procedures and evaluation metric. We then present the comparative qualitative results against the state-of-the-art models in Sec.~\ref{exp:qual_cmp}, followed by more visual results of the ablation studies in Sec.~\ref{exp:abl}. 
In Sec.~\ref{exp:granularity} and Sec.~\ref{exp:concepts}, we provide further analysis on the visual results with different granularity, as well as different concepts, respectively. 
Sec.~\ref{exp:fairness} provides a fairness comparison on an unseen dataset.
In Sec.~\ref{exp:app}, we show the superiority of our foundation model over Stable Diffusion pretrained model on two downstream applications.
Finally, we provide additional visualizations of our model in Sec.~\ref{exp:more_samples}.

\subsection{Experiment Setting}
\label{exp:setting}
Our foundation models CosmicMan contains two versions: CosmicMan-SD based on SD-1.5~\cite{sd}, and CosmicMan-SDXL based on SDXL~\cite{sdxl}.
The DDPMScheduler~\cite{ddpm} and EulerAncestralDiscreteScheduler are used for training and testing, respectively. 
The HOLA loss only affects the cross-attention block at 16$\times$ down-sampling rate.
In CosmicMan-SDXL, the multi-aspect training strategy \cite{sdxl} in combination with an offset-noise level of 0.05 is utilized to train our foundation model, and the weight of HOLA loss is set to 1.
In CosmicMan-SD, the weight of HOLA loss is set to 0.001 to avoid over-saturation problem.
During the training process, a dropout strategy is applied with a probability of 0.1 to all attributes located in front of each object, as well as to the region description phrases, with the notable exception of the phrase ``Overall''. Additionally, the photo type, derived from the keypoints, is systematically positioned at the forefront of the dense description.
Besides, we also use the quality tuning technique \cite{emu} on both models to further enhance the final visually appealing. We finetune on 500 extremely high-quality human images with an offset-noise level of 0.05 for about 1000 iterations. 

\begin{figure*}[!t]
    \centering
    \includegraphics[width=\textwidth]{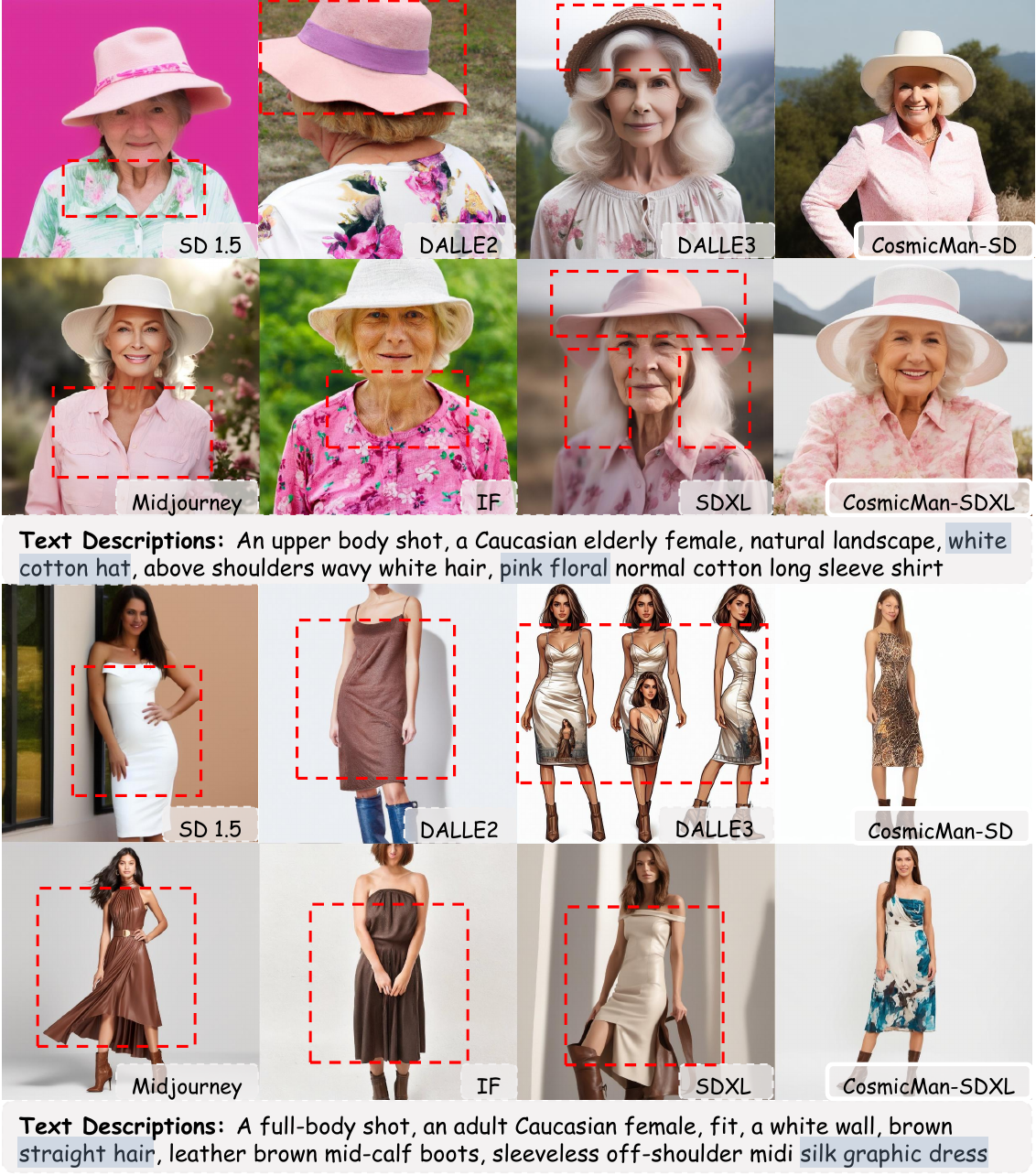}
    \caption{\textbf{Comparison with State-of-the-art Models.} From left to right, the top row features the results of SD 1.5, DALLE2, DALLE3, and CosmicMan-SD. The bottom row presents the results of Midjourney 5.2, DeepFloyd-IF, SDXL, and CosmicMan-SDXL.}
    \label{fig:sota1}
    \vspace{20pt}
\end{figure*}

\begin{figure*}[t]
    \centering
    \includegraphics[width=\textwidth]{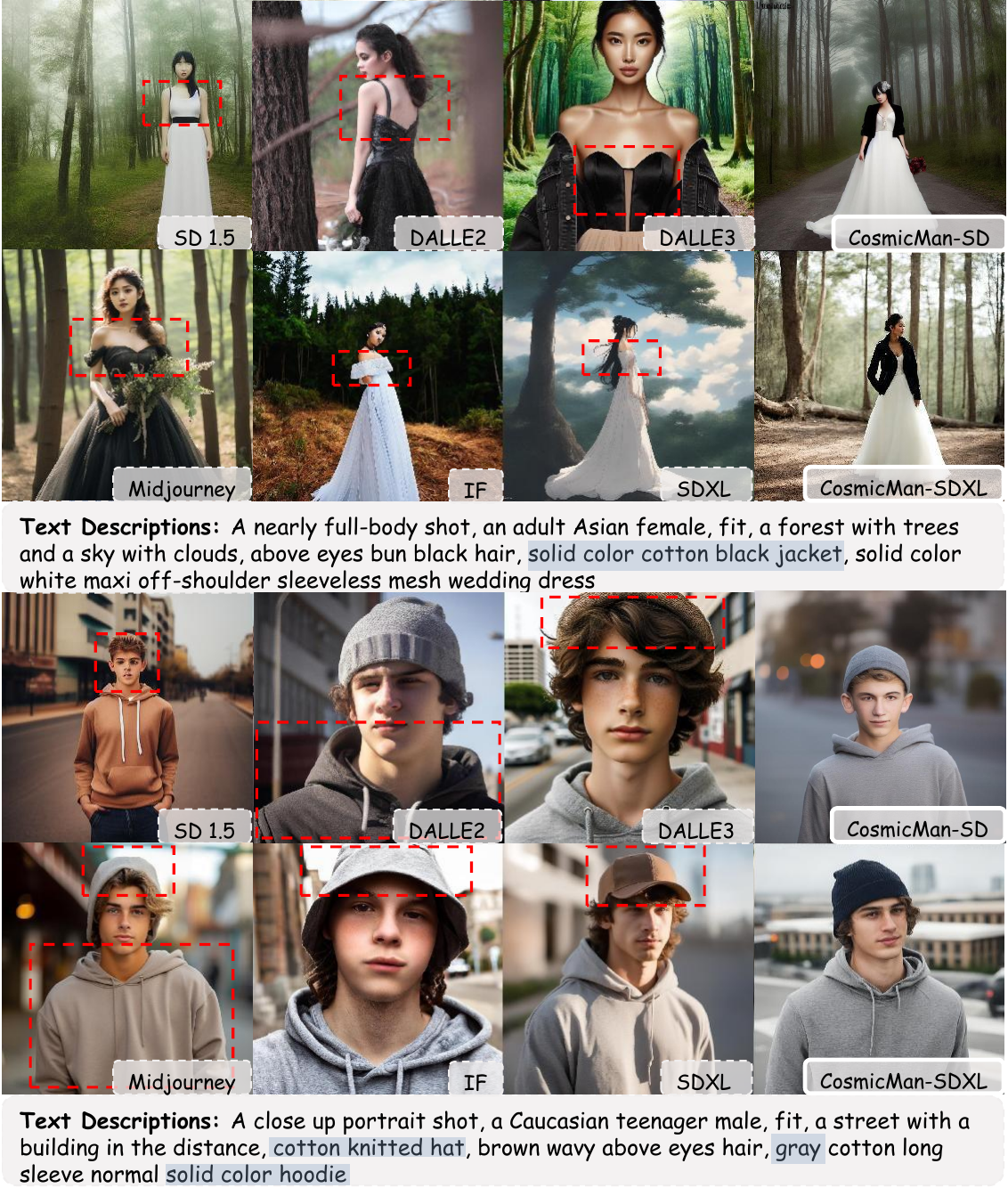}
    \caption{\textbf{Comparison with State-of-the-art Models.} From left to right, the top row features the results of SD 1.5, DALLE2, DALLE3 and CosmicMan-SD. The bottom row presents the results of Midjourney 5.2, DeepFloyd-IF, SDXL, and CosmicMan-SDXL.}
    \label{fig:sota2}
\end{figure*}

\begin{figure*}[t]
    \centering
    
    \vspace{20pt}
    \includegraphics[width=1\textwidth]{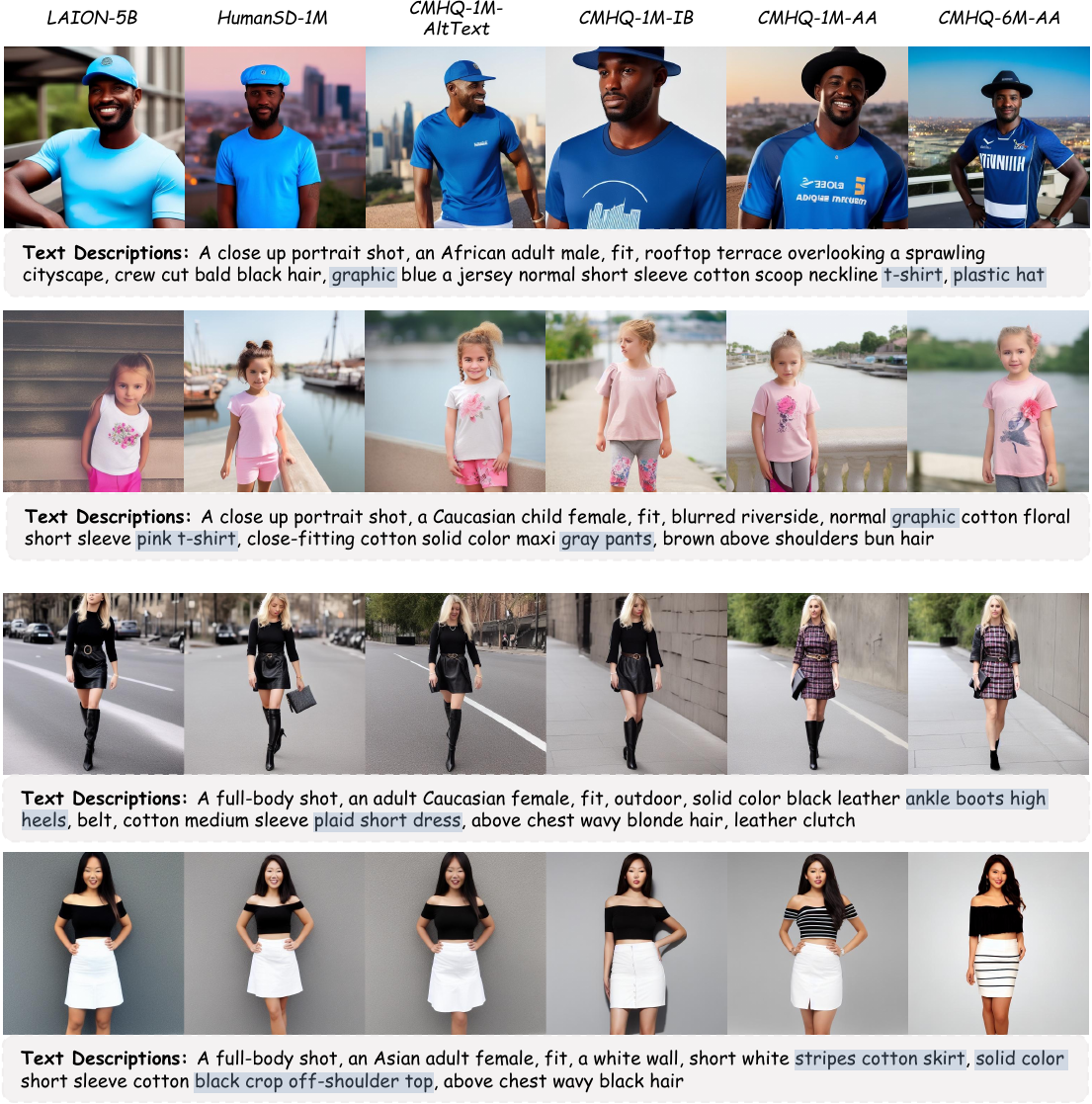}
    \caption{\textbf{Ablation on Training Data.} 
     ``AltText'' refers to Web Alternative Text, ``IB'' denotes the image descriptions generated by the pretrained InstructBLIP model, ``AA'' corresponds to captions produced by Annotate Anyone, and ``CMHQ'' refers to the CosmicMan-HQ.}
    \label{fig:train_data}
    \vspace{2.5cm}
\end{figure*}

\subsection{Fine-grained Text-Image Alignment Metric} \label{exp:metric}
We introduce \textbf{Semantic Acc}, a novel text-image alignment metric specifically designed for dense concepts in human images. Firstly, it adopts an atomized approach as other metrics for fine-grained text-image alignment~\cite{dsg, compbench}, breaking down descriptions into discrete questions to minimize coupling. This atomization is effectively implemented using a predefined question dictionary for finite human descriptions, thereby avoiding the additional errors introduced by generating questions through Large Language Models. 
For heightened precision, we fine-tuned a Vision-Language Model specifically to answer the atomized questions, rather than relying on pretrained models. Specifically, we ask ``yes'' or ``no'' questions for each atomic attribute using InstructBLIP. For example, for a generated image with input description of ``plaid short sleeve t-shirt'', we would ask three questions sequentially: ``Does the person wear a t-shirt?'', ``Does the t-shirt have a plaid pattern?'', and ``Is the t-shirt short sleeve?''. The answer to all three questions is ``yes''. To prevent overfitting, attributes of the same hierarchy are randomly selected from CosmicMan-HQ 1.0 to generate ``no'' answer questions during training. The Semantic Acc scores are calculated by dividing the number of correct answers by the total number of questions. To further analyze text-image alignment for humans across various dimensions, we categorize Semantic Acc into three groups: \textbf{Acc\textsubscript{obj}} for object types, \textbf{Acc\textsubscript{tex}} for texture attributes, and \textbf{Acc\textsubscript{shape}} for shape attributes, as indexed in Tab.~\ref{tab:detailed_questions}.



    

\subsection{Qualitative Comparison to SOTA}
\label{exp:qual_cmp}
We focus on comparing our two foundational models with state-of-the-art models such as SD-1.5, DALLE2, DALLE3, Midjourney, DeepFloyd-IF, and SDXL, as depicted in Fig.~\ref{fig:sota1} and Fig.~\ref{fig:sota2}. While SD-1.5 shows the highest CLIPScore among these contenders, as highlighted in the main paper, its visual performance indicates subpar image quality and text-image alignment. Other models are able to generally produce human images that are consistent with detailed descriptions, though they sometimes omit or misinterpret certain concepts, which are highlighted by red dotted boxes in figures. For instance, in the first caption of Fig.~\ref{fig:sota2}, which depicts a white wedding dress and a black jacket, these models successfully generate the dress but fail to capture the concept of the black jacket. Regarding CosmicMan-SD, which possesses the same architecture as SD-1.5, it exhibits better text-image alignment. CosmicMan-SDXL excels in both image quality and text-image alignment for dense concepts.

\subsection{Qualitative Ablation Studies}
\label{exp:abl}
In this section, We provide qualitative  results based on our setting to show the effectiveness of each part proposed in Daring.

\noindent
\textbf{Ablation on Training Dataset.}
In evaluating the efficacy of data sources, it is observed that the outputs derived from the model trained on CosmicMan-HQ 1.0 dataset exhibit a closer resemblance to authentic imagery and a higher text-image alignment when compared with those generated by models trained on LAION-5B and HumanSD. This comparative analysis is exemplified in Fig.~\ref{fig:train_data}, where the first row illustrates a notable discrepancy. The LAION-5B image is characterized by a slight overexposure, while the representation of clothing material in the HumanSD image deviates from a cotton-like texture.

Regarding the impact of data scaling on model performance, the training on the CosmicMan-HQ-6M dataset demonstrates enhanced precision in the text-image alignment for dense concepts. As evidenced in Fig.~\ref{fig:train_data}, the model trained with CosmicMan-HQ-6M more accurately captures fine-grained attributes, such as ankle boots (illustrated in the third row) and a striped skirt (depicted in the fourth row), in contrast to its CosmicMan-HQ-1M counterpart.

Furthermore, the superiority of annotation quality in the Annotate Anyone becomes evident when contrasted with the pretrained IB model. This is particularly noticeable in the generation of more accurate and refined annotations, leading to significant improvements in the visual representation of items such as a graphic pink t-shirt and gray pants, as showcased in the second row of Fig.~\ref{fig:train_data}.

\begin{figure}[t]
    \centering
    \includegraphics[width=1\linewidth]{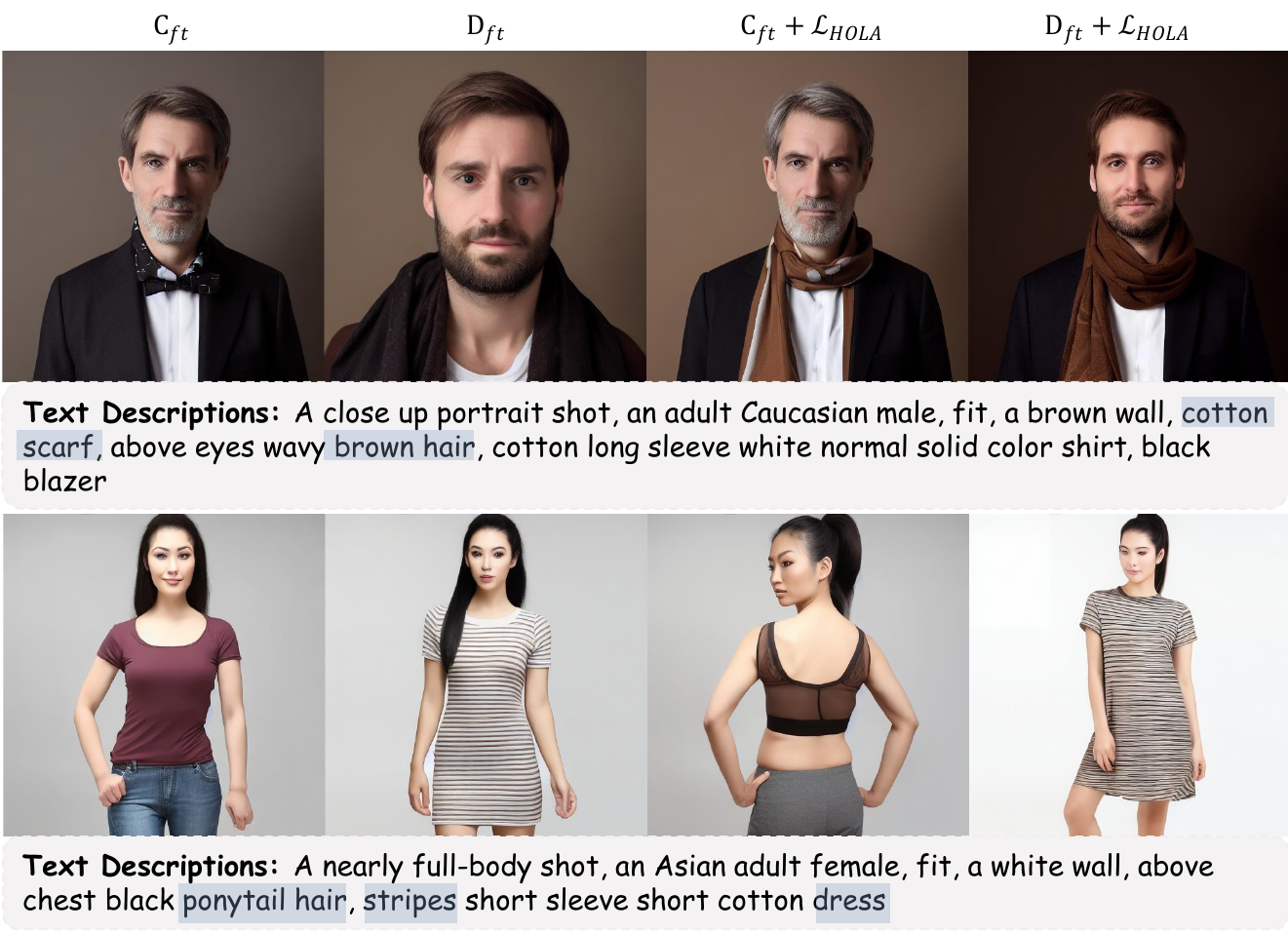}
    \caption{\textbf{Ablation on HOLA Loss and Data Discretization.} ${D}_{ft}$ and ${C}_{ft}$ denote fine-tuning and testing models on the continuous and decomposed text space. $\mathrm{\mathcal{L}_{HOLA}}$ represents the human body and outfit-guided loss for alignment. The blue backgrounds are utilized to accentuate the content of the generated pairs. This approach is consistently applied throughout the following images to maintain uniformity and enhance visual clarity. 
    }
    \label{fig:text_sapce}
\end{figure}

\begin{figure*}
    \centering
    \includegraphics[width=1\linewidth]{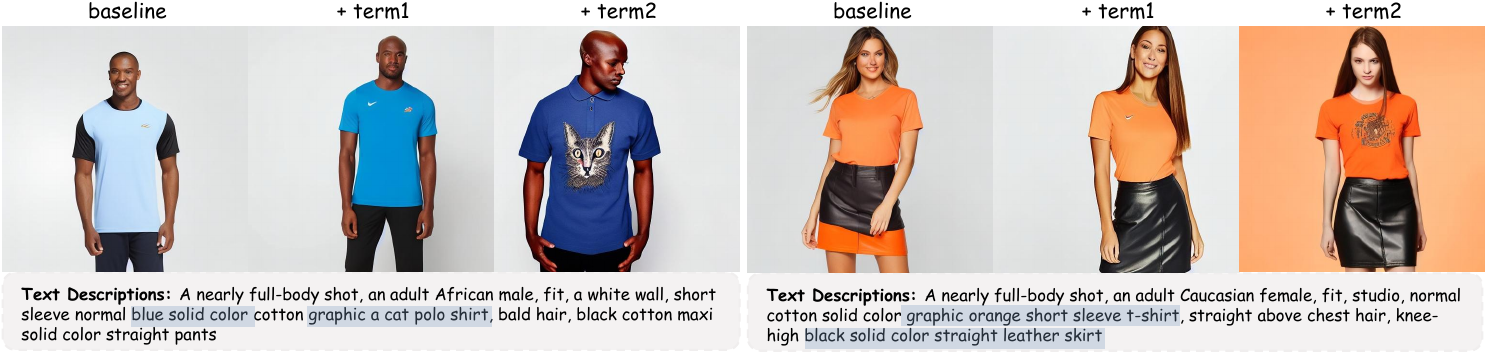}
    \caption{\textbf{Ablation on Terms of HOLA Loss.} Baseline means only fine-tuning the model on CosmicMan-HQ 1.0. Term 1 refers to the first term of HOLA loss, which works under the guidance of human body structure. Term 2 refers to the second term of HOLA loss, which helps reduce the ambiguities of outfit-level descriptions.}
    \label{fig:loss-term}
\end{figure*}

\noindent
\textbf{Ablation on Training Strategy.}
We delineate the impact of Data Discretization for Humans and the incorporation of HOLA loss on the accuracy of the generated images.
Fig.~\ref{fig:text_sapce} elucidates this effect by contrasting results obtained under decomposed and continuous text spaces, underscoring the significance of Data Discretization for Humans. The model $D_{ft}$ demonstrates a heightened ability to accurately generate images of a cotton scarf (as depicted in the first row) and a striped dress (shown in the second row), compared to its counterpart, $C_{ft}$. This observation underscores the premise that a discrete text space is more adept at handling complex, dense concepts in our dataset. Nevertheless, the absence of constraints on key $K$ in the attention block occasionally leads to instances of attribute misalignment.
In an effort to address this, the integration of HOLA loss with Data Discretization for Humans is explored. This combination, represented as $D_{ft}+$$\mathrm{\mathcal{L}_{HOLA}}$, culminates in a further enhancement of semantic alignment. The resulting images depict a more realistic representation of a cotton scarf and accurately colored hair in the first row, and a more precisely rendered striped dress and ponytail hair in the second row.

In an effort to further dissect the influence of HOLA loss, a comparative analysis focusing on the individual contributions of each term within the loss function is conducted. This is depicted in Fig.~\ref{fig:loss-term}, where the differential impacts of these terms are illustrated.
Specifically, Term 1 is observed to facilitate the improved generation of attributes that are characteristic of larger regions. Examples of this can be seen in the enhanced depiction of a blue shirt and a black skirt on the left and right images, respectively. However, it is noted that Term 1 is less effective in accurately rendering attributes that are localized. For instance, the polo shirt on the left side is more akin to a t-shirt in its representation. This discrepancy can be attributed to the nuanced differences, such as the localized collar shape, which distinguishes a polo shirt from a t-shirt.
By composing Term 2, the model exhibits a heightened proficiency in generating attributes pertinent to localized regions. This is evident in the more accurate portrayal of the cat pattern and the polo shirt on the left, as well as the distinct pattern of the t-shirt on the right.

\begin{figure*}[!t]
    \centering
    \includegraphics[width=0.95\textwidth]{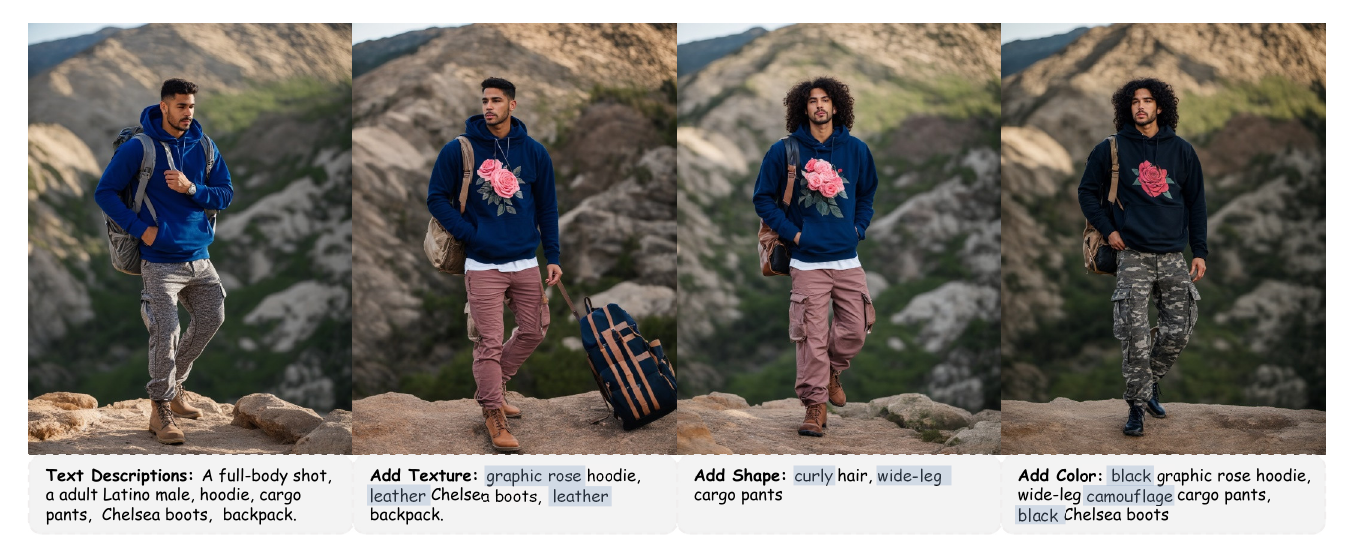}
    \caption{\textbf{Qualitative Results of CosmicMan-SDXL with Increasing Granularity.} A simple description is initially provided, from which the granularity is progressively increased, emphasizing texture, shape, and color, leading to the generation of the following human images.}
    \label{fig:increasing_granularity}
\end{figure*}

\subsection{Qualitative Results of Different Granularity}
\label{exp:granularity}
To illustrate CosmicMan's ability in processing descriptions of varying granularity, we offer a visualization in Fig.~\ref{fig:increasing_granularity} that features captions with progressively increasing detail. Initially, we start with a simple caption specifying different outfit types to generate the base human image. Subsequently, we incrementally enrich the description by adding details about texture, shape, and color. The step-by-step result reveals our model's capability to produce high-quality human images that remain faithfully aligned with increasingly complex and dense concepts.

\subsection{Qualitative Results of Different Concepts}
\label{exp:concepts}
We demonstrate the versatility of CosmicMan-SDXL in handling varying fine-grained descriptions, as depicted in Fig.~\ref{fig:diff_description}. We only modify certain elements of the descriptions while keeping other components consistent. In the first row, we alter the descriptions of outfit shape, such as the sleeve length in the target region, resulting in visible changes, yet the overall spatial layout is maintained. Next, we explore different textures and colors, like fur floral dress, showcasing the model's ability to incorporate unique dense concepts. Lastly, when modifying the outfit type, significant changes are observed in the human structure and scene layout.

\subsection{Quantitative Comparison on Unseen Testset}
\label{exp:fairness}
We follow the experimental settings and provide an additional zero-shot evaluation on an unseen human subset, which contains $2048$ images filtered from the MS-COCO 2014 validation subset~\cite{mscoco}. As shown in Tab.~\ref{tab:quan_cmp}, our models outperform all other methods in terms of image quality~(FID). Regarding text-image alignment, we compare CLIPScore instead of Semantic Acc since the captions are in free-form text format. Its results consistently align with those from our original test set.

\begin{table}[H]
    \caption{\textbf{Quantitative Comparison on Unseen Human Testset.} We conduct zero-shot evaluation on MS-COCO 2014 validation subset. ``CM-SD'' and ``CM-SDXL'' are short for CosmicMan-SD and CosmicMan-SDXL. } 
    \centering
    \renewcommand{\arraystretch}{1}
    \resizebox{1\linewidth}{!}{
    \begin{tabular}{ccccccc}
    \toprule
      Metrics  & SD 1.5  & SD 2.0 & SDXL & DF-IF & CM-SD & CM-SDXL \\
    \midrule
     FID & 50.45 & 75.71  &  50.16 &  53.38 &  49.34 & 45.90 \\
     CLIP &  27.23 & 23.58  &  27.37 & 26.96 & 27.23  & 25.56 \\
    Speed~(ms/f) & 43.66 & 55.99 & 56.23 & 203.43 & 43.53 & 56.32 \\
    \bottomrule
    \end{tabular}
    }
    \label{tab:quan_cmp}
\end{table}




\subsection{Applications} 
\label{exp:app}
In this subsection, we provide more qualitative results on two downstream applications: 2D human editing and 3D human reconstruction to show the effectiveness of our specialized foundation model.

\noindent
\textbf{2D Human Editing.}
We compare our ComsicMan-SDXL with SDXL pretrained model using T2I-Adapter~\cite{t2i-adapter}. We use skeleton maps extracted by Openpose~\cite{openpose} as guidance to generate portraits with specified poses. The first and second rows in Fig.~\ref{fig:supp_app1} show that our results exhibit more accurate pose control. Besides, SDXL tends to generate semantic inconsistency images with a hazy background, while our method generates more realistic and semantic-consistent images.

\noindent
\textbf{3D Human Reconstruction.}
We further compare our ComsicMan-SD with SD-1.5 pretrained model based Magic123~\cite{magic123} for 3D human reconstruction. 
The first two examples in Fig.~\ref{fig:supp_app2} show the ability on Image-to-3D, where both reference image and prompt are used as input. The first example shows that our model can maintain the hat shape of the girl in each view, demonstrating that our model possesses better multi-view consistency. The second example shows Stable Diffusion pretrained model tends to generate results with vague and odd human shapes (the 3rd and 6th images of SD in second example). In contrast, our results obtain more accurate human body and face geometric shapes. The third example in Fig.~\ref{fig:supp_app2} shows the ability on Text-to-3D, where the single prompt is initially input to CosmicMan-SD or Stable Diffusion pretrained model to generate the reference image, and then sent to Magic123. It can be seen that our results are significantly superior to Stable Diffusion results on both text-image consistency and geometric shape.

\begin{figure*}[tp]
    \centering
    \includegraphics[width=0.825\textwidth]{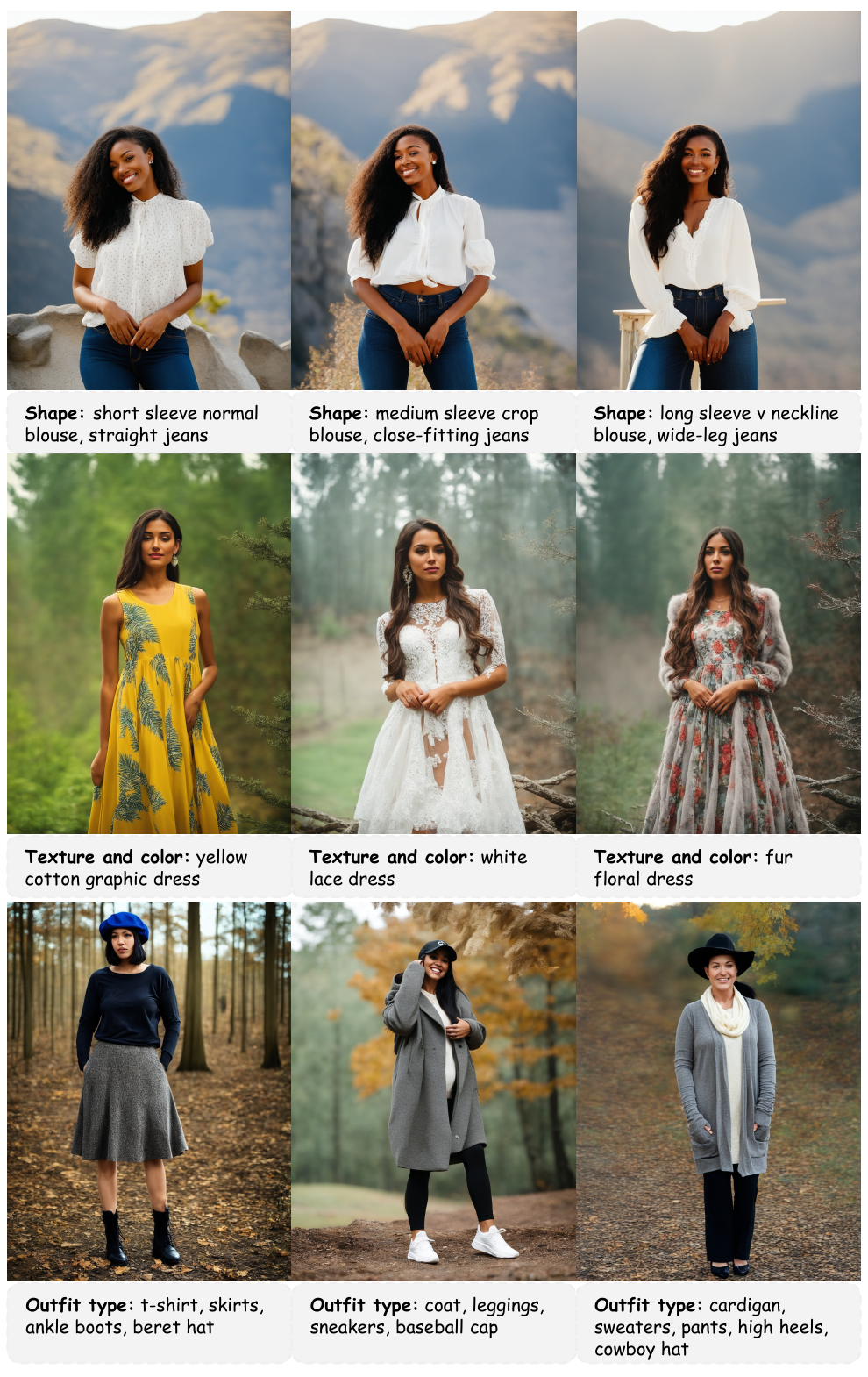}
    \vspace{-0.35cm}
    \caption{\textbf{Qualitative Results of CosmicMan-SDXL with Different Descriptions on Shape, Texture, and Outfit Types.}}
    \label{fig:diff_description}
\end{figure*}

\begin{figure*}[tp]
    \centering
    \includegraphics[width=0.825\textwidth]{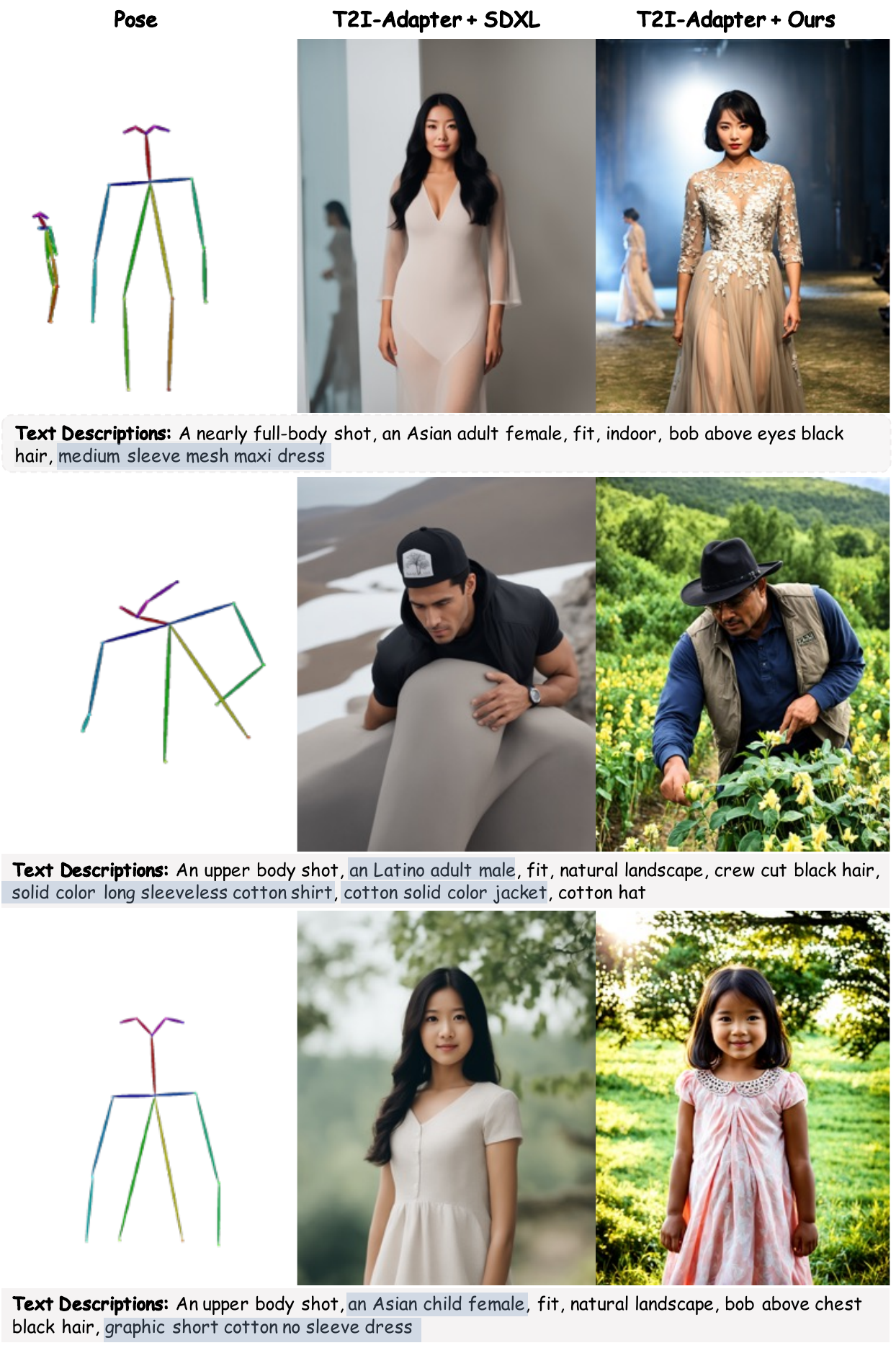}
    \caption{\textbf{Visualization of 2D Human Editing.} We compare our CosmicMan-SDXL with SDXL pretrained model based on T2I-Adapter~\cite{t2i-adapter}. We highlight the areas of text-image inconsistency of SDXL results in ``Text Descriptions'' using blue background.}
    \label{fig:supp_app1}
\end{figure*}

\begin{figure*}
    \centering
    \includegraphics[width=1\textwidth]{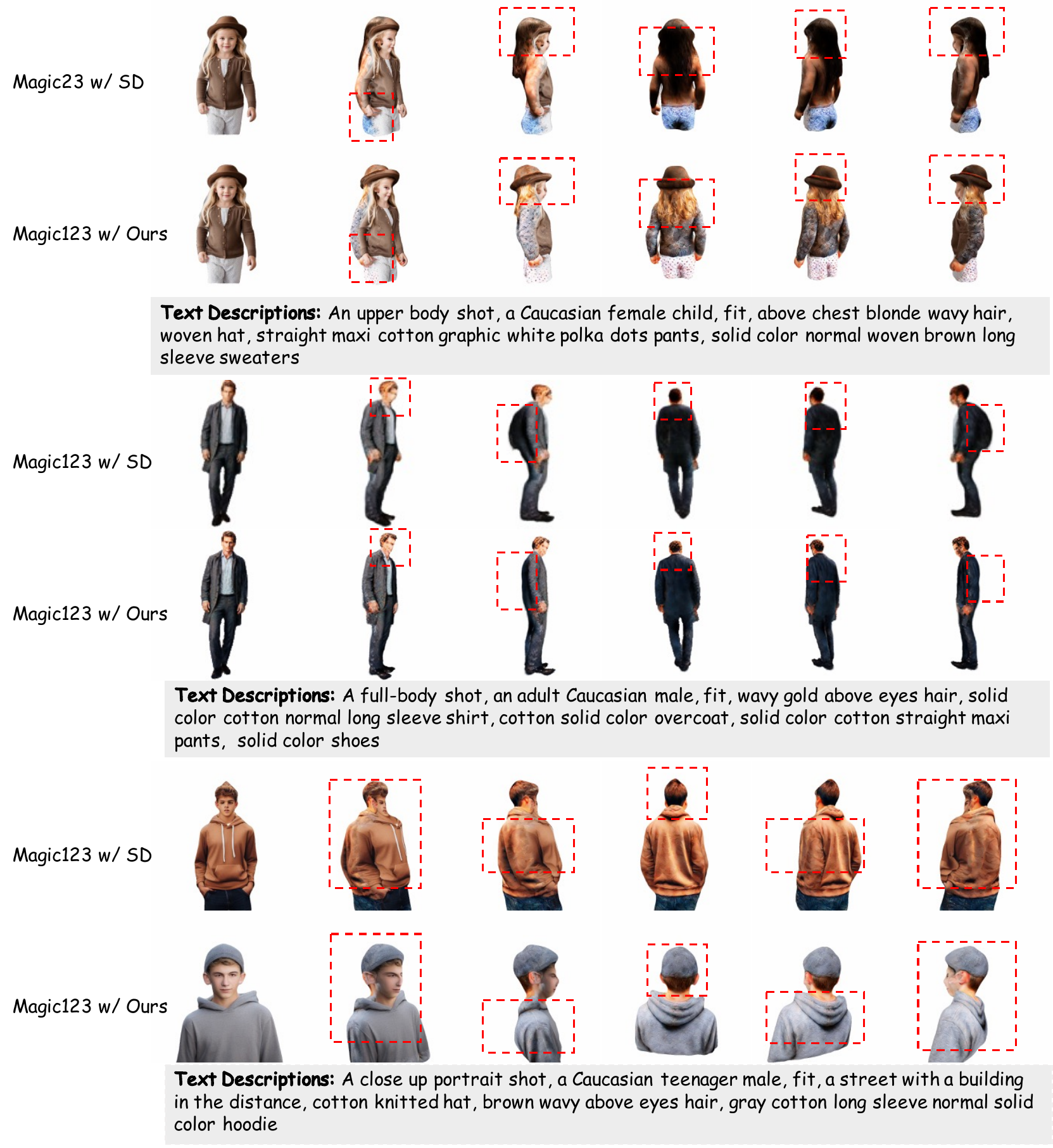}
    \caption{\textbf{Visualization of 3D Human Reconstruction.} We compare our CosmicMan-SD with Stable Diffusion pretrained model based on Magic123~\cite{magic123}. The first and second examples show the ability of Image-to-3D. The third example shows the ability of Text-to-3D. CosmicMan-SD can generate results with better multi-view consistency and geometric shapes. The first image in each row is the reference image. Note that the reference images in the fifth and sixth are generated by CosmicMan-SD and SD-1.5 respectively. The red box highlights the improvement of our foundation model compared to Stable Diffusion.}
    \label{fig:supp_app2}
\end{figure*}

\subsection{More Generated Samples of CosmicMan}
\label{exp:more_samples}
In Fig.~\ref{fig:sampled_results_0} $\sim$ Fig.~\ref{fig:sampled_results_5}, we present additional generated samples showcasing a variety of captions and aspect ratios. Please note that some images have been cropped for improved typesetting.

\begin{figure*}
    \centering
    \includegraphics[width=0.9\textwidth]{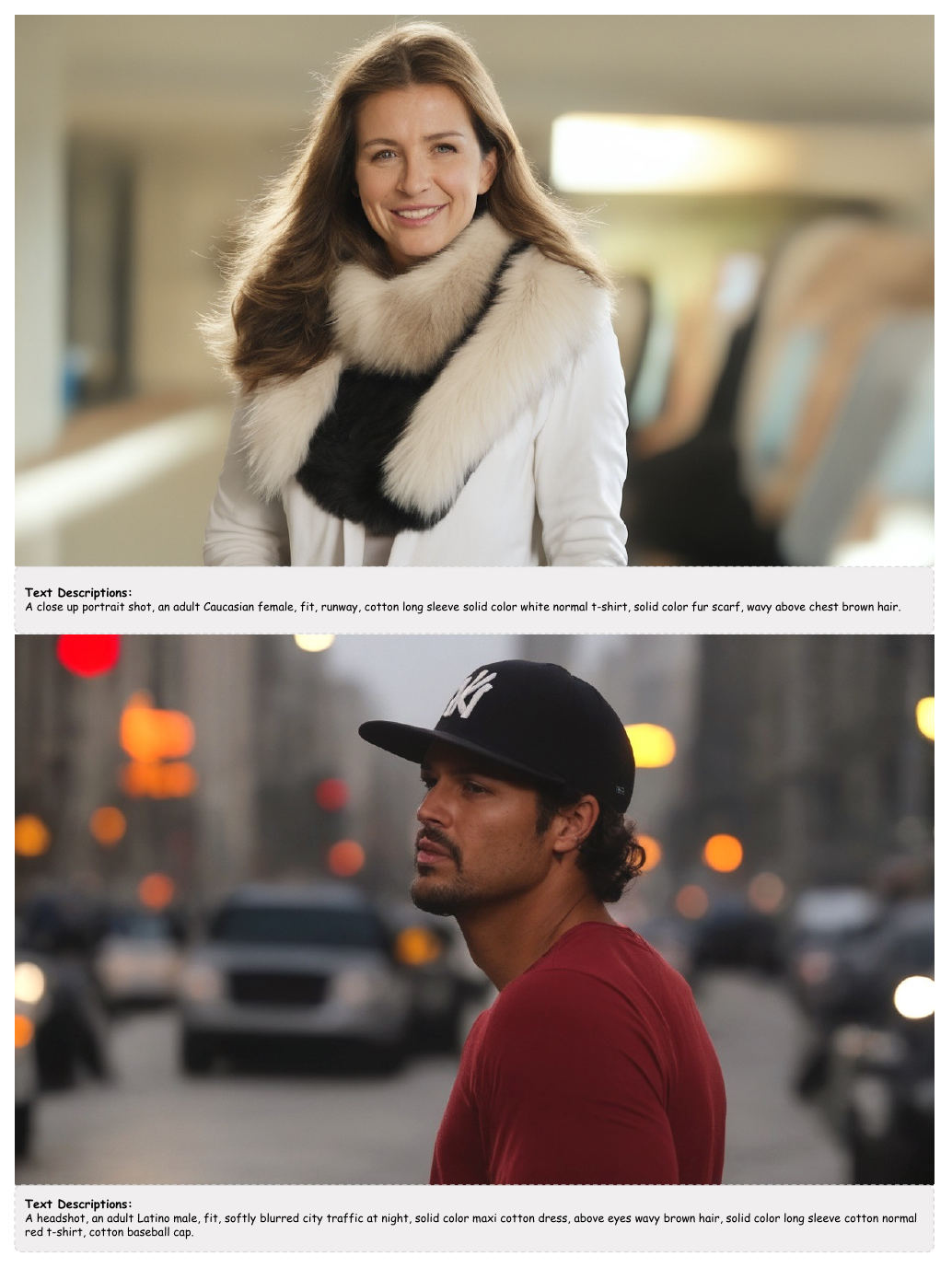}
    \caption{\textbf{More Generated Samples of CosmicMan.}}
    \label{fig:sampled_results_0}
\end{figure*}

\begin{figure*}
    \centering
    \includegraphics[width=0.9\textwidth]{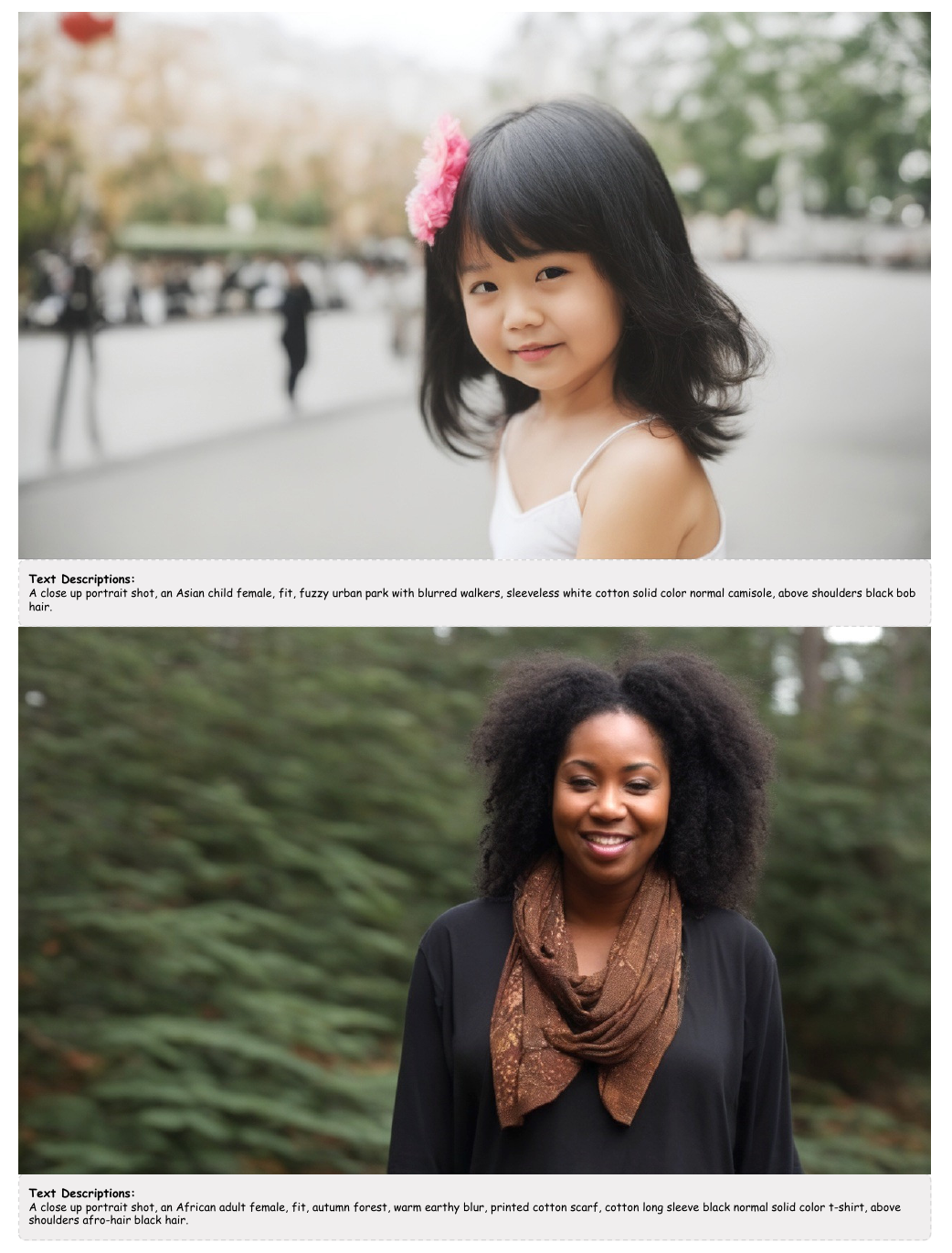}
    \caption{\textbf{More Generated Samples of CosmicMan.}}
    \label{fig:sampled_results_1}
\end{figure*}

\begin{figure*}
    \centering
    \includegraphics[width=1\textwidth]{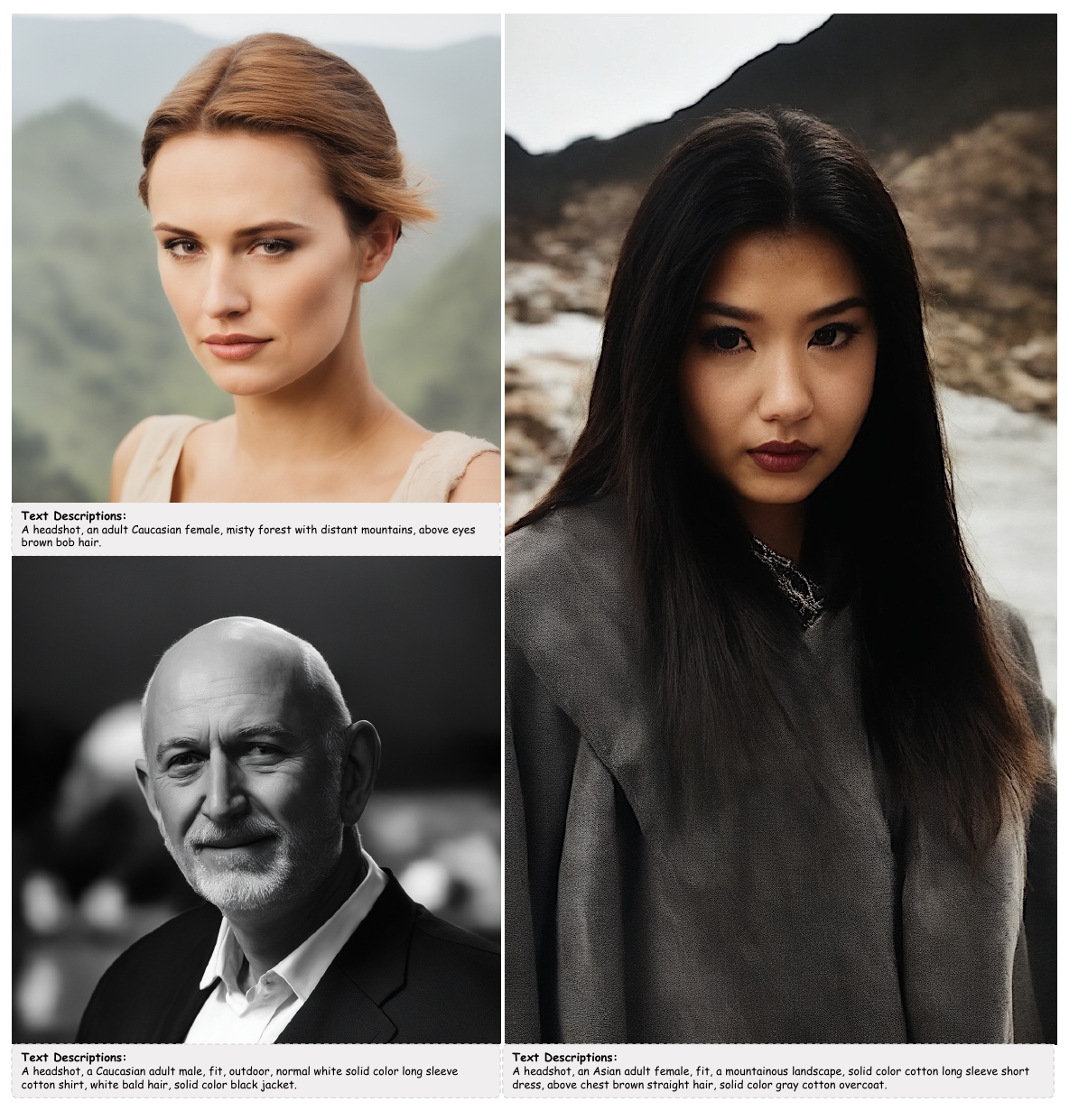}
    \caption{\textbf{More Generated samples of CosmicMan.}}
    \label{fig:sampled_results_2}
\end{figure*}

\begin{figure*}
    \centering
    \includegraphics[width=1\textwidth]{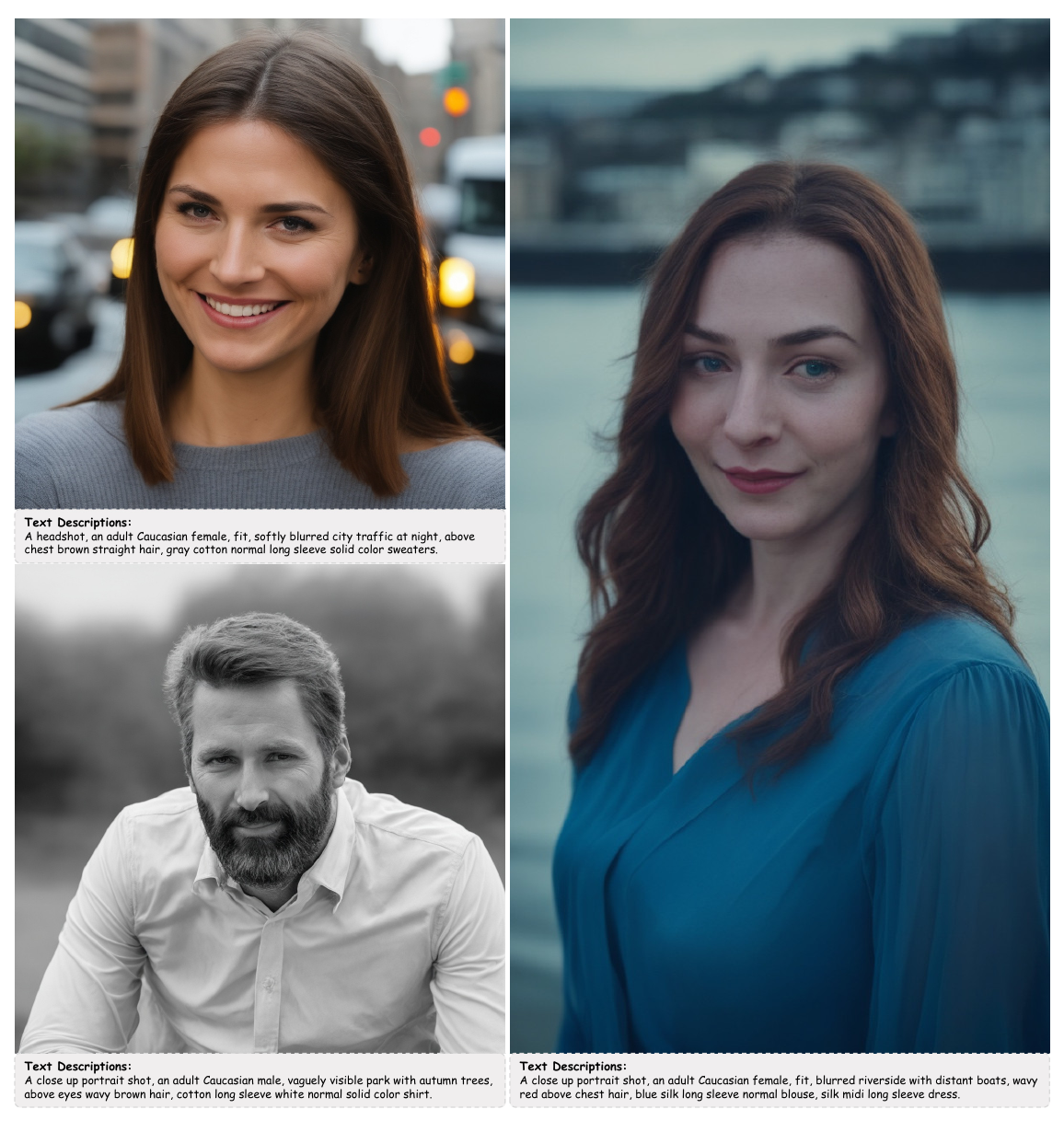}
    \caption{\textbf{More Generated Samples of CosmicMan.}}
    \label{fig:sampled_results_3}
\end{figure*}

\begin{figure*}
    \centering
    \includegraphics[width=0.825\textwidth]{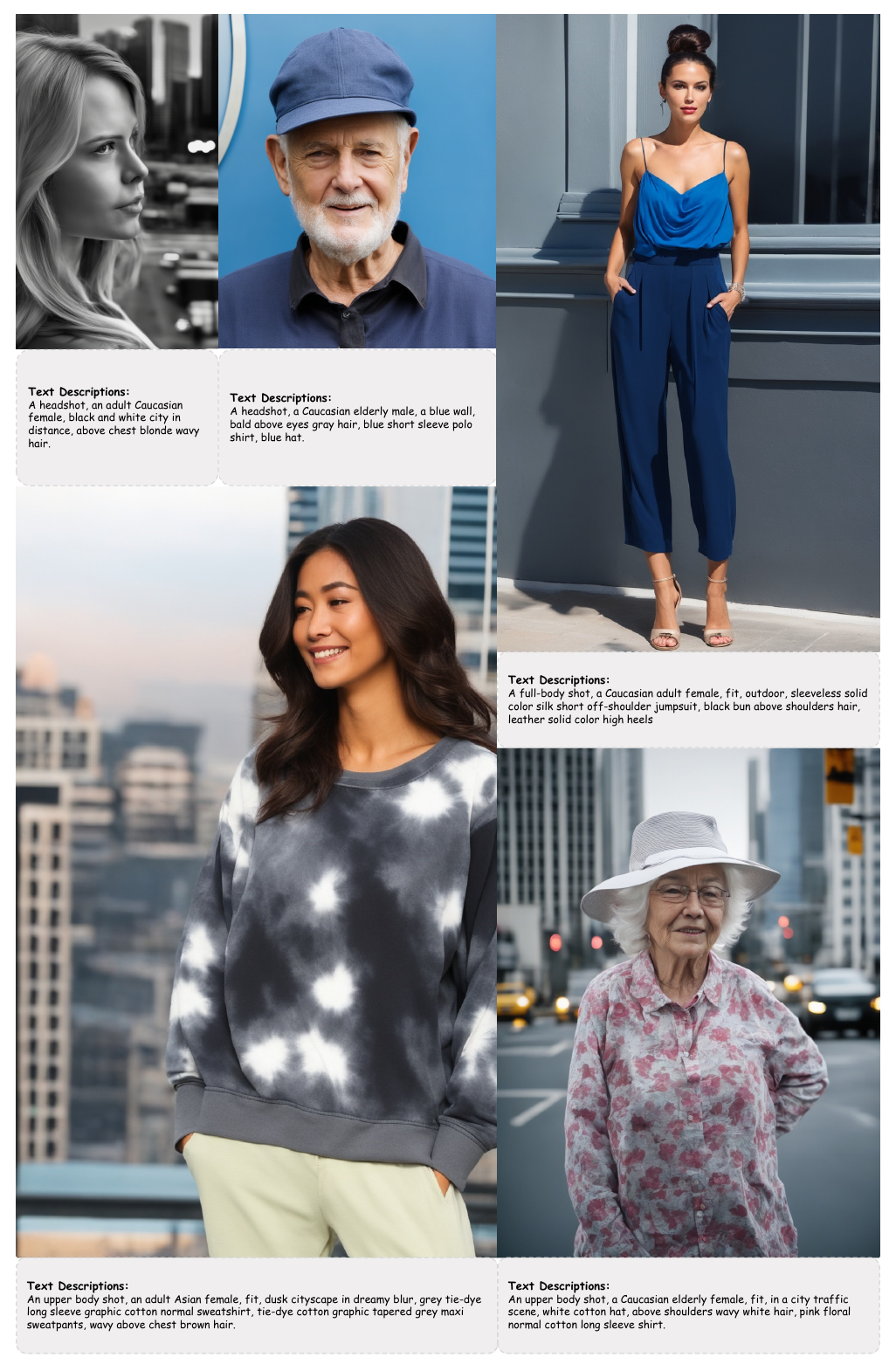}
    \caption{\textbf{More Generated Samples of CosmicMan.}}
    \label{fig:sampled_results_4}
\end{figure*}

\begin{figure*}
    \centering
    \includegraphics[width=0.825\textwidth]{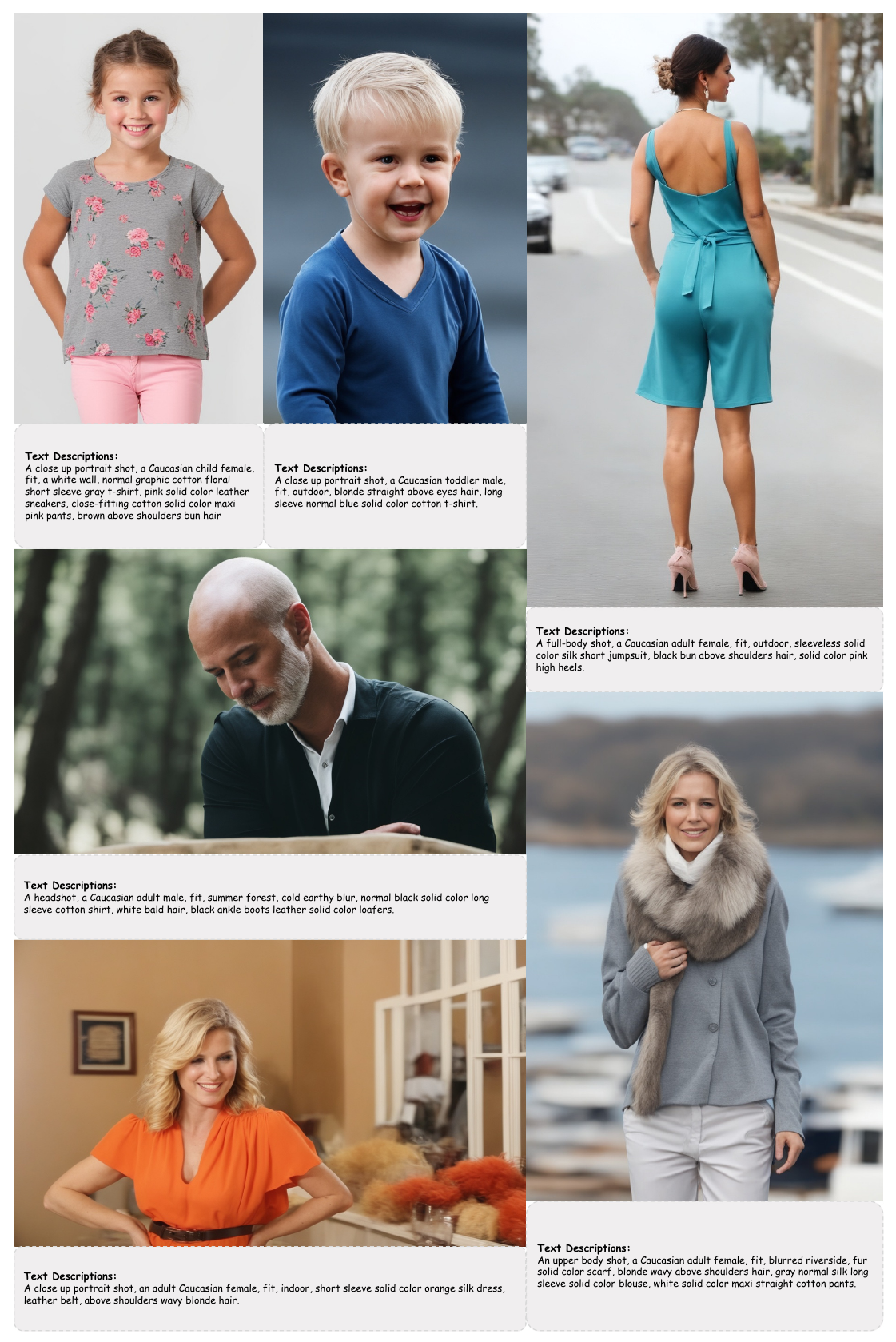}
    \caption{\textbf{More Generated Samples of CosmicMan.}}
    \label{fig:sampled_results_5}
\end{figure*}

\clearpage
\clearpage


\end{document}